\pdfoutput=1
\documentclass[11pt]{article}
\usepackage[preprint]{acl}
\usepackage{times}
\usepackage{multirow}
\usepackage{arydshln}
\usepackage{balance}
\usepackage{graphicx}
\usepackage{xspace}
\usepackage{booktabs} 
\usepackage{tcolorbox}
\usepackage[hypcap=false]{caption}
\usepackage{latexsym}
\usepackage[T1]{fontenc}
\usepackage[utf8]{inputenc}
\usepackage{microtype}
\usepackage{amsmath}
\usepackage{inconsolata}
\usepackage{graphicx}
\usepackage{xcolor}
\usepackage{makecell} 
\usepackage{array}
\usepackage{algorithmic}
\usepackage{colortbl}
\usepackage{xcolor}
\usepackage{algorithm}
\usepackage{tabularx}

\definecolor{cFuji}{RGB}{31,119,180}
\definecolor{cBen}{RGB}{255,127,14}
\definecolor{cSnowdon}{RGB}{44,160,44}
\definecolor{cKosciuszko}{RGB}{214,39,40}
\definecolor{cToubkal}{RGB}{148,103,189}

\newcommand{\ctag}[2]{\textcolor{#1}{\textbf{[#2]}}}
\newcommand{\cspan}[2]{\textcolor{#1}{\textbf{#2}}}

\definecolor{pos}{RGB}{215,48,39}   % red
\definecolor{neg}{RGB}{69,117,180}  % blue
\definecolor{udqbg}{RGB}{219,209,233}
\definecolor{plotorange}{RGB}{255,127,14}
\newcommand{\good}[1]{\textcolor{pos}{#1}}
\newcommand{\bad}[1]{\textcolor{neg}{#1}}
\usepackage[table]{xcolor}
\usepackage{comment}

\definecolor{mygreen}{RGB}{0, 153, 0}
\newcommand{\method}{\textsc{Diverge}\xspace}
\usepackage[table]{xcolor}
\usepackage{tikz}
\usepackage{tcolorbox}
\usetikzlibrary{tikzmark,calc}

\newcommand{\tabcaption}[1]{\vspace*{-3mm}\caption{#1}\vspace*{-4mm}}
\newcommand{\figcaption}[1]{\vspace*{-3mm}\caption{#1}\vspace*{-5mm}}
\newcommand{\moveup}{\vspace*{-2mm}}
\newcommand{\moveups}{\vspace*{-1mm}}
\newcommand{\ignore}[1]{}
\usepackage[utf8]{inputenc}
\usepackage{hyphenat}
\usepackage{xspace}
\usepackage{amsmath}
\usepackage{amsfonts}
\usepackage{hyperref}
\usepackage{url}
\usepackage{booktabs}
\usepackage{multirow}
\usepackage{makecell}
\usepackage{minibox}
\usepackage{bbm}
\usepackage{graphicx}
\usepackage{balance}
\usepackage{mathtools}
\usepackage{color}
\usepackage{marvosym}
\usepackage{ifthen}
\usepackage{textcomp}
\usepackage{enumitem}
\usepackage{verbatim}
\usepackage{algorithm}
\usepackage{algorithmic}
\usepackage{numprint}
\usepackage{balance}

\newcommand{\chatoDisplayMode}[1]{#1}

\definecolor{MyRed}{rgb}{0.6,0.0,0.0} 
\definecolor{MyBlack}{rgb}{0.1,0.1,0.1} 
\newcommand{\inred}[1]{{\color{MyRed}\sf\textbf{\textsc{#1}}}}
\newcommand{\frameit}[2]{
  \begin{center}
  {\color{MyRed}
  \framebox[.9\columnwidth][l]{
    \begin{minipage}{.85\columnwidth}
    \inred{#1}: {\sf\color{MyBlack}#2}
    \end{minipage}
  }\\
  }
  \end{center}
}

\newcommand{\note}[2][]{\chatoDisplayMode{\def\@tmpsig{#1}\frameit{{\Pointinghand} Note}{#2\ifx \@tmpsig \@empty \else \mbox{ --\em #1}\fi}}}
\newcommand{\abbrevStyle}[1]{#1}
\newcommand{\cf}{\abbrevStyle{cf.}\xspace}

\newcommand{\xhdr}[1]{\vspace{1.7mm}\noindent{{\bf #1.}}}

\newcommand{\textcite}[1]{\citeauthor{#1} \shortcite{#1}}

\title{DIVERGE: Diversity-Enhanced Retrieval-Augmented Generation
for Open-Ended Information Seeking }

\author{
  Tianyi Hu\textsuperscript{1} \quad
  Niket Tandon\textsuperscript{2} \quad
  Akhil Arora\textsuperscript{1} \\
  \textsuperscript{1} Aarhus University \quad
  \textsuperscript{2} Microsoft Research \\
  \texttt{\{tenney.hu, akhil.arora\}@cs.au.dk}
}

\begin{document}
\maketitle
\begin{abstract}
%Existing retrieval-augmented generation (RAG) systems typically assume that each query has a single correct answer, overlooking common information-seeking scenarios with multiple plausible answers, where diversity is crucial to prevent responses from collapsing into a single dominant perspective and to support creativity, fairness, and inclusive information access.
%Our analysis reveals a key limitation of standard RAG systems: they underuse diverse retrieved contexts, so increasing retrieval diversity alone does not necessarily produce diverse generations.
%To address this limitation, we propose \textbf{\method}\footnote{Code is available at: \href{https://anonymous.4open.science/r/Diverge/}{link}}, a plug-and-play agentic RAG framework that promotes diverse viewpoints through reflection-guided generation and memory-augmented iterative refinement while preserving answer quality.
%We introduce novel human-aligned metrics for evaluating the diversity–quality trade-off in open-ended questions. Experiments across multiple real-world open-ended QA datasets and backbone LLMs show that \method achieves the best trade-off, outperforming competitive baselines by substantially improving diversity while preserving quality. Our results reveal a systematic limitation of current LLM-based systems and show that explicit diversity modeling can mitigate it.
Existing retrieval-augmented generation (RAG) systems often assume that each query has a single correct answer. This assumption overlooks open-ended information-seeking scenarios where multiple plausible answers are valuable, and where diversity is important for creativity, fairness, and inclusive access to information.
We show that standard RAG systems fail to fully use diverse retrieved contexts: simply increasing retrieval diversity does not necessarily lead to diverse generations. To address this limitation, we propose \textbf{\method}\footnote{Code is available at  \href{https://github.com/au-clan/diverge}{this repo}}, a plug-and-play agentic RAG framework that improves the diversity--quality trade-off through iterative, reflection-guided exploration of diverse viewpoints and diversity-aware retrieval support.
We further introduce evaluation metrics for characterizing the diversity--quality trade-off in open-ended question answering. Experiments across multiple real-world datasets and backbone LLMs show that \method achieves the best trade-off among competitive baselines, increasing diversity by $\sim2\times$ without noticeable quality degradation. These results reveal a systematic limitation of current RAGs and show the value of explicit diversity modeling.
\end{abstract}
\section{Introduction}
%Motivation
Retrieval-augmented generation (RAG)~\cite{lewis2020retrieval} enhances LLMs' ability to ground responses in external knowledge and improve quality in knowledge-intensive tasks.
However, most prior works~\cite{zhang2025faithfulrag,yang2018hotpotqa,yu2024evaluation} are built upon the hypothesis that \textit{each question has a single, clearly defined factual answer}. While enabling factual grounding, it overlooks the fact that real-world information-seeking needs are often \textit{open-ended} where a query may have \textit{multiple plausible answers}~\cite{wikimedia2030, arora_nav, jiang2025artificial}, 
, as users' cultures~\cite{hershcovich-etal-2022-challenges}, values~\cite{solaiman2021process}, and personal preferences~\cite{sorensen2024roadmap} can shape which answers they find relevant, useful, or preferable.

\begin{figure*}[t]
    \moveup
    \moveups
        \centering
        \includegraphics[
            width=0.82\linewidth,
            trim=0cm 1cm 0cm 0cm,
            clip
        ]{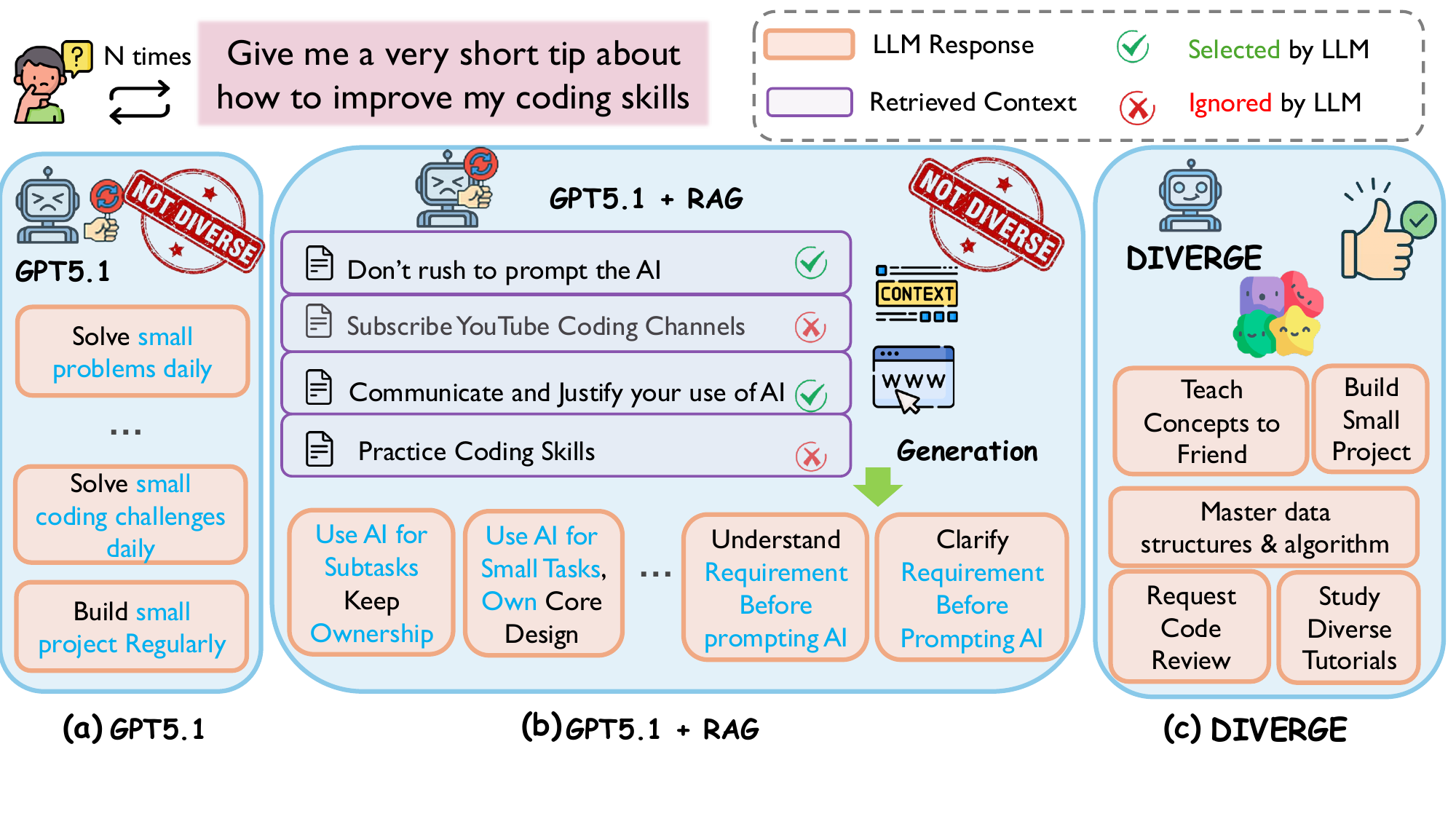}
    \moveups
    \moveups
\figcaption{\emph{Illustrative example} of an open-ended information-seeking query. (a) LLMs exhibit homogenized (\bad{blue}) outputs, (b) standard RAG still produces repetitive responses even when the retrieved contexts contain diverse evidence. In contrast, (c) \method generates diverse outputs while maintaining high answer quality.}
    \label{fig:example}
\end{figure*}

In open-ended settings, response diversity is essential for representing diverse viewpoints fairly and inclusively, and for preventing homogenized LLM outputs from narrowing human creativity~\cite{zhang2025noveltybench}. 
At the same time, ensuring high answer quality while promoting diversity remains a key challenge~\cite{lanchantin2025diverse,shypula2025evaluating}.
Prior work suggests that post-training can cause current closed-book LLMs to concentrate probability mass on a narrower set of high-reward modes~\cite{karan2025reasoning,pandey2024brain}. As a result, these models often fail to preserve output diversity, leading to homogenized generation regimes.
Although result diversification has been studied in information retrieval (IR)~\cite{khan2013scalable}, simply integrating diverse IR techniques into RAG pipelines does not guarantee diverse generations, leaving it unclear whether RAG systems can overcome the inherent homogenization tendencies of current LLMs.

\xhdr{Challenges} As a concrete illustration of challenges (\S~\ref{sec: preliminary}), Fig.~\ref{fig:example} shows a key limitation of RAG for open-ended questions, where diverse retrieved contexts fail to translate into diverse outputs due to LLM homogenization. 
Our investigation (\S~\ref{sec: main_res}) further shows that simple strategies are insufficient: increasing retrieval diversity alone does not yield more diverse generations, while state-of-the-art (SoTA) prompt-based baselines~\cite {zhang2025verbalized} for diversity enhancement achieve only limited gains in diversity at the cost of substantial quality degradation in open-ended information-seeking.

Motivated by these observations, we systematically analyze the diversity challenges faced by existing RAG systems (see \S~\ref{sec:challenges}) and identify three key issues: (C1) \textbf{Single-Answer Bias}, where individual RAG generations prioritize overconfident answers and overlook alternative information; (C2) \textbf{Missing Diversity Preservation}, where the lack of mechanisms for tracking previously explored viewpoints leads to highly similar responses across generations; and (C3) \textbf{Practical Compatibility}, as many prior methods require token-level logits and are thus incompatible with frontier LLMs.

\xhdr{Present work} To address these challenges, we propose \textbf{\method} (\underline{\textbf{Div}}ersity-\underline{\textbf{E}}nhanced \underline{\textbf{R}}etrieval-Augmented \underline{\textbf{Ge}}neration), a novel plug-and-play agentic RAG framework explicitly designed to address the diversity–quality trade-off in real-world open-ended information-seeking (\S~\ref{sec:diverge}),  equipped with novel components that explicitly promote diversity while preserving answer quality.
Specifically, \method addresses the three challenges by:
(i) mitigating the single-answer bias (\textit{C1}) through explicit reflection on uncovered viewpoints; (ii) enabling long-horizon diversity preservation while maintaining answer quality (\textit{C2}) via an iterative RAG process with lightweight memory and evidence-grounded generation; (iii) avoiding reliance on token-level logits (\textit{C3}), thereby ensuring compatibility with arbitrary LLM backbones, including closed-source frontier models.  

Most existing RAG evaluation metrics~\cite{es2024ragas} rely on predefined ground-truth answers and therefore do not scale to open-ended settings. Moreover, existing diversity–quality trade-off evaluations largely focus on creative tasks~\cite{lanchantin2025diverse}, leaving a gap for information-seeking scenarios. To facilitate this, 
% evaluation, 
we introduce a novel set of metrics. 
% For \textit{diversity}, 
To capture both high-level diversity and the diversity across multiple viewpoints within a single response, we consider two complementary dimensions:
\textit{semantic diversity}, which measures diversity of the overall response, and
\textit{coverage diversity}, which decomposes a response into a set of atomic viewpoints and measures diversity across them.
 For \textit{quality}, given the open-ended nature of the task and aligned with convention~\cite{badshah2024reference,xu2025ask}, we adopt an LLM-as-a-judge paradigm. 
Finally, to enable an intuitive comparison of the trade-off, we propose a \textit{Unified Diversity–Quality Harmonic Score} (\textit{Unified Score}).

We empirically validate \method on two complex real-world open-ended benchmarks, \emph{Infinity-Chat}~\cite{jiang2025artificial} and \emph{IssueBench}~\cite{rottger2025issuebench}. \method achieves the \textbf{highest} \textit{Unified Score} across all methods, delivering  ~$\mathbf{2\times}$ improvements in both semantic and coverage diversity over direct prompting, with only negligible impact on quality.  We further demonstrate the effectiveness of our framework through analyses (\S~\ref{sec: ana}). Our contributions can be summarized as follows: 

\begin{itemize}[leftmargin=*, itemsep=0pt, topsep=0pt]
\item We identify a commonly overlooked limitation of RAG systems: in open-ended information seeking, they suffer from knowledge collapse and underutilize diverse contexts, while existing prompt-based methods often degrade quality.

\item We propose metrics for open-ended information seeking that capture the diversity--quality trade-off and enable intuitive, systematic comparison.

\item We introduce \method, a plug-and-play diversity-enhanced RAG compatible with frontier LLMs, and show that it achieves the best diversity--quality trade-off in real-world settings.
\end{itemize}

\section{Related Work}
\xhdr{Homogeneity of LLMs}
Recent studies~\cite{jiang2025artificial,zhang2025noveltybench} show that LLMs generate significantly less diverse outputs than human authors. This homogeneity has raised broad concerns, including social and cultural biases from dominant perspectives~\citep{rottger2025issuebench}, epistemic collapse~\citep{wright2025epistemic}, failures in customizable AI systems~\citep{zhang2025noveltybench}, and homogenized human thinking under exposure to LLM-generated content~\cite{jiang2025artificial}. Prior work suggests that this homogeneity is driven by post-training that sharpens models’ output probability distributions~\citep{lanchantin2025diverse}, and preference data also systematically reward more typical responses, further biasing models toward less diverse outputs~\cite{zhang2025verbalized}. Although RAG can access diverse knowledge~\cite{wright2025epistemic}, it remains constrained by the homogenized generation behavior of its backbone LLMs, which tend to favor deterministic outputs over alternatives~\cite{zharzhavsky2026bluff}. As a result, whether RAG can produce diverse responses in open-ended settings remains largely unexplored.

\begin{figure*}[t!]
    \centering
    \moveup
    \includegraphics[
        width=\linewidth,
        trim=0 120 0 0,
        clip
    ]{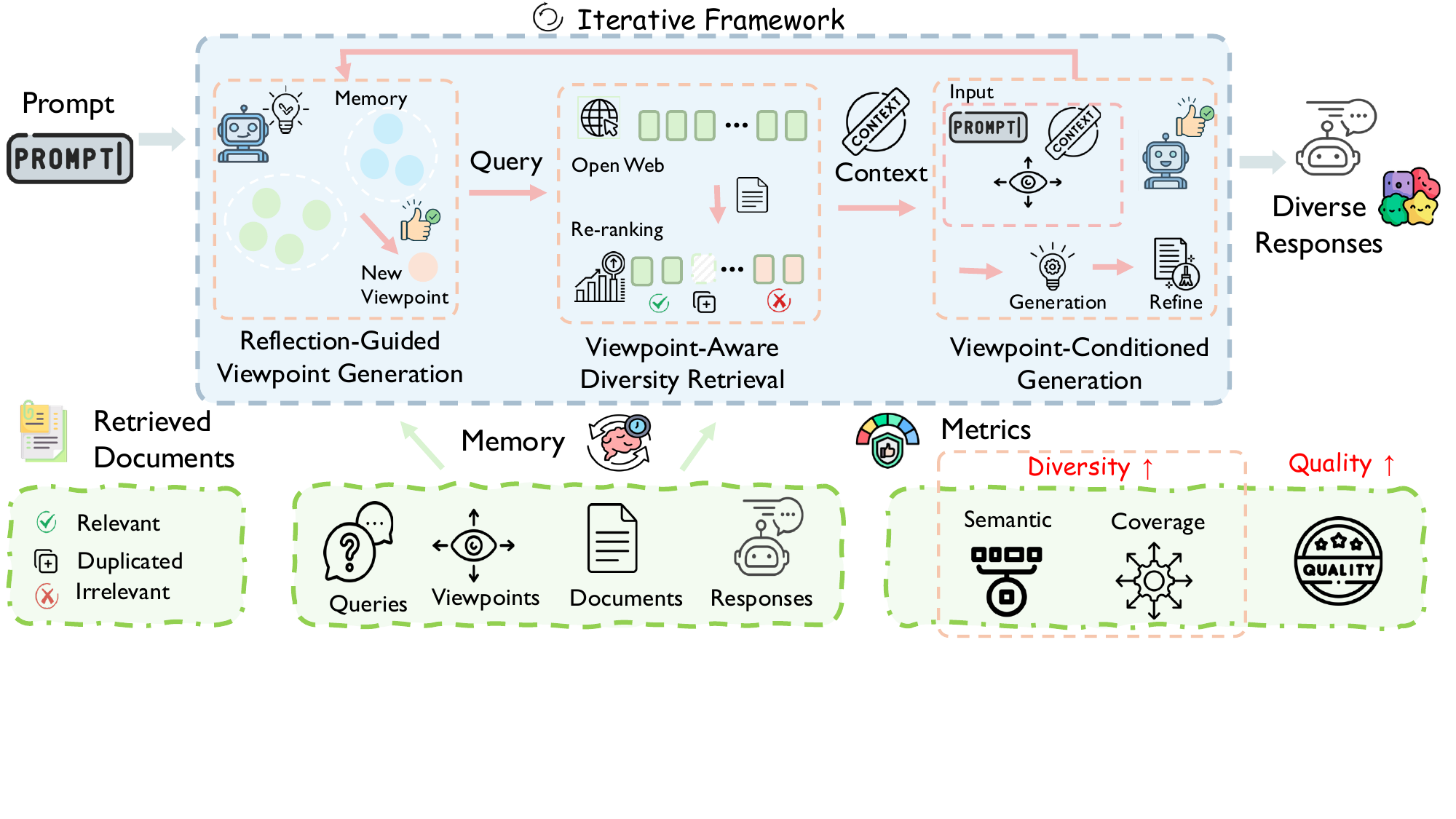}
    \moveup
    \moveup
    \figcaption{Overview of \method, a plug-and-play RAG framework for open-ended QA that promotes diverse viewpoints via reflection-guided viewpoint generation and viewpoint-conditioned RAG, with broad LLM compatibility.}
    \label{fig:method}

\end{figure*}

\xhdr{Increasing Output Diversity}
A wide range of test-time strategies has been proposed to improve output diversity, primarily by adjusting decoding hyperparameters such as temperature, top-$p$, top-$k$~\citep{shi2024thorough}, and min-$p$~\cite{nguyen2024turning}. However, these methods provide limited gains when models exhibit collapsed output distributions~\cite{jiang2025artificial}, and many closed-source LLMs (e.g., GPT-5) no longer expose such controls.
Another line improves diversity by retraining LLMs with diversity-aware alignment objectives, such as \textit{DivPO}~\cite{lanchantin2025diverse}. While effective, these methods are resource-intensive and do not apply to closed-source frontier LLMs. Prompt-based approaches could also explore more diverse outputs~\cite{shur2024growing,zhang2025verbalized}; however, they often increase diversity at the cost of answer quality (\cf \S~\ref{sec: main_res}).
Together, these limitations motivate our work, which improves diversity without compromising answer quality and applies to any base model.

\xhdr{Diversity in IR and RAG}
In IR, diversity has long been pursued to cover a broader range of user preferences via techniques such as query rewriting and re-ranking~\citep{mohankumar2021diversity}. However, IR systems stop at returning ranked document lists, leaving users to inspect the evidence themselves~\cite{li2025towards}. RAG reduces this burden by grounding responses directly in retrieved evidence.
In RAG, prior work on diversity has primarily operated on the input side, either retrieving comprehensive contexts to support multi-hop reasoning~\cite{wang2025diversity,rezaei2025vendi,khan2026df} or building diverse training data~\cite{liu2025rag}. 
DeepResearch-style systems~\cite{xu2025comprehensive,li2025muisqa} follow the same pattern, aggregating broad evidence to converge on a single consolidated answer. Across these lines, all works retain the \textit{single-answer assumption} and aim for correctness. 
%Another line is to improve the richness of RAG outputs~\cite{wang2025richrag}; while not strictly tied to a single explicit answer, it is still confined to a narrow set of pre-collected reference answers.
Even when the target is not strictly a single answer, the acceptable answer space is still typically constrained to a narrow set of predefined references~\cite{wang2025richrag}. However, open-ended information-seeking queries may admit an unbounded space of plausible answers.
Other related studies either focus on narrow domains, such as cross-cultural recipe adaptation~\cite{hu2025culinary}, or only briefly discuss in knowledge collapse~\cite{wright2025epistemic}.  As a result, the design and evaluation of RAG for output diversity in real-world open-ended settings remain largely unexplored and is the focus of our work.

\section{Preliminary}
\label{sec: preliminary}
\subsection{Task formulation}
Following prior work~\cite{jiang2025artificial,zhang2025noveltybench,wright2025epistemic}, we formalize the task as: 
Given an arbitrary diversity metric $\mathcal{D}$, quality metric $\mathcal{Q}$, and an arbitrary model hyperparameters configuration $c$, the task takes as input a set of open-ended queries
\( \mathbb{Q} = \{ q^1,\ldots, q^N \} \). For each query \( q^i \), the model produces a set of \( K \) responses, we denote this set by \(\mathcal{A}^i_c = \{ a^i_{c,1}, \ldots, a^i_{c, K} \}\). The objective is to produce responses that collectively exhibit both high diversity and quality. Quality can be easily assessed by averaging each output. So the primary question is how to define diversity.

\citeauthor{kirk2023understanding} (\citeyear{kirk2023understanding}) propose two paradigms for measuring diversity: \textit{across-input}, which measures variation from a global perspective across inputs, and \textit{per-input} diversity, which captures diversity only among multiple outputs from the same input. We adopt the \textit{per-input} paradigm, as our focus is on assessing the diversity of responses generated for the same open-ended query. So it is defined as:

\moveup
\moveup
\begin{equation*}
\mathrm{Diversity}_{\mathcal{D}}(c)
:= \frac{1}{N} \sum_{i=1}^{N} \mathcal{D}\!\left(\mathcal{A}^i_c\right).
\end{equation*}
\moveup
\moveup
\moveup

\subsection{Diversity Challenges for RAG}
\label{sec:challenges}

We identify three key challenges that make it difficult for existing RAG systems to support diverse output in open-ended settings.

\xhdr{C1: Single-Answer Bias} At the level of individual generations, existing RAG systems are optimized to produce reliable and accurate answers under the single-answer assumption~\cite{zhang2025rag}, which typically leads to low uncertainty~\cite{zharzhavsky2026bluff}, even less variation than LLMs that are already highly homogenized~\cite{soudani2025uncertainty}. This causes each response to prioritize a narrow, high-confidence subset of contexts and ignore alternative yet plausible information~\cite{hu2025culinary}.

\vspace{-4pt}

\xhdr{C2: Missing Diversity Preservation Across Generations} While C1 concerns individual responses, diversity also degrades across multiple generations due to the lack of explicit mechanisms for tracking previously explored viewpoints. Open-ended questions typically involve many perspectives~\cite{jiang2025artificial} distributed across different sources, yet existing systems repeatedly generate similar viewpoints, leading to redundant outputs.
\vspace{-4pt}

\xhdr{C3: Practical Compatibility} Beyond these conceptual challenges, most existing test-time diversity-enhancing approaches~\cite{nguyen2024turning,shi2024thorough} rely on token-level logits, which remain unavailable in most closed-source frontier LLMs~\cite{hiranandani2025logits}. Furthermore, an emerging trend among these models is to prohibit the use of decoding hyperparameters such as temperature~\cite{openai2025temperature}, as observed in recent models such as GPT-5 series, further limiting real-world applicability.

\section{\method}
\label{sec:diverge}

\xhdr{Overview} We introduce \method, a plug-and-play agentic RAG framework that introduces viewpoints as an explicit intermediate abstraction for controlling diversity in open-ended generation. At each iteration, \method mitigates single-answer bias (\emph{C1}) by casting diversity exploration as a test-time reasoning process, reflecting over previous responses to identify uncovered perspectives, while reducing diversity collapse (\emph{C2}) through a lightweight memory that carries explored viewpoints and retrieved evidence across iterations, enabling the model to avoid redundancy. Unlike naive combinations of retrieval with prompt-based diversity methods, which degrade quality without meaningful diversity gains (§\ref{sec: main_res}), \method explicitly structures both retrieval and generation around viewpoints to preserve quality. Since \method operates as a plug-and-play pipeline without requiring token-level logits, it can be applied to arbitrary LLM backbones, including closed-source models (\emph{C3}). Figure~\ref{fig:method} illustrates the framework, and Algorithm~\ref{alg:divrag} and Appendix~\ref{app: method} provide the full procedure.

\xhdr{Reflection-Guided Viewpoint Generation} 
Prior research~\cite{wang2022self} suggests that multiple viable internal reasoning trajectories can coexist within LLMs, and that appropriate prompting can steer models toward different directions~\cite{zhuo2024prosa}. 
Moreover, mechanistic analyses indicate that multiple latent features coexist within models and can be selectively activated~\cite{anthropic2023monosemantic}. Inspired by these insights, we conceptualize these latent features as \textit{viewpoints} and use them as a core abstraction in our framework. \method first summarizes the initial RAG response and then iteratively reflects on prior outputs to maintain a set of existing viewpoints. At each iteration, the LLM identifies a new, insufficiently covered viewpoint based on those previously explored, thereby avoiding repeated generation. This reflection-guided process promotes the exploration of alternative perspectives and mitigates the tendency to repeatedly generate responses from a single dominant stance.

\xhdr{Viewpoint-Aware Diversity Retrieval}
Viewpoints provide diverse perspectives but are hypothetical and may lack factual grounding. To ground each viewpoint, \method uses the LLM to generate a viewpoint-conditioned query $q_t$, retrieves evidence from the open web, and applies diversity-aware re-ranking. The re-ranker jointly considers relevance to $q_t$, non-redundancy with previously retrieved evidence, and diversity among documents selected in the current iteration. Specifically, we extend MMR~\cite{carbonell1998use} with an iteration-aware ranking score:

\moveup
\moveup
{\normalsize
\begin{equation*}
\label{eq:div_rerank_t}
\begin{aligned}
s_t(d) \;=\;
&\;\alpha \cdot \mathrm{Rel}(d, q_t)
\;-\;
\beta \cdot \max_{h \in \mathcal{M}_{<t}} \mathrm{Sim}(d, h) \\[-0.4em]
&\;-\;
(1 - \alpha) \cdot \max_{s \in \mathcal{S}_t} \mathrm{Sim}(d, s) .
\end{aligned}
\end{equation*}}
\moveup
\moveup
\moveup

Here, $t$ is the current iteration, $\mathcal{M}_{<t}$ stores contexts retrieved in previous iterations, and $\mathcal{S}_t$ contains documents already selected in the current iteration. $\mathrm{Rel}$ and $\mathrm{Sim}$ denote relevance and embedding-based cosine similarity, while $\alpha$ and $\beta$ control the relevance--diversity trade-off.

\xhdr{Viewpoint-Conditioned Generation}
Even with a novel viewpoint and supporting evidence, generation may still fail to fully address the user query, either by weakly connecting to the original question or by omitting essential information. To address this, \method introduces a viewpoint-conditioned generation and refinement process. The model first generates a response conditioned on the original query, the target viewpoint, and the retrieved evidence, then it is refined to align with the original query, strengthening logical coherence, and preventing deviation while preserving completeness.

Together, these components equip \method to explore diverse viewpoints while maintaining high output quality. Results are presented in ~\S~\ref{sec: main_res}.

\section{Evaluating Diversity-Quality Trade-offs}
\label{sec: metric}

Our evaluation uses both existing and newly introduced metrics to assess \textit{diversity} (\S~\ref{sec: div}), \textit{quality} (\S~\ref{sec: quality}), and further introduces \textit{unified score} (\S~\ref{sec: unified}) for a summary of the diversity--quality trade-off in our task (examples in Appendix Figure ~\ref{fig:metric_example}).

\subsection{Diversity Metrics}
\label{sec: div}
For diversity, we consider two complementary dimensions. 
First, we adopt \textit{semantic diversity}~\cite{jiang2025artificial,guo2024benchmarking}, a widely used metric that captures overall variation in meaning across responses. 
Second, we introduce \textit{coverage diversity}, a task-oriented metric tailored to open-ended information-seeking, which decomposes long responses into atomic viewpoints and measures how broadly different plausible answers are covered.

\xhdr{Semantic Diversity}
Semantic diversity~\cite{guo2024benchmarking} measures variation in meaning among generated responses.
For each query, we compute semantic diversity as the average pairwise normalized cosine distance between answer embeddings. The resulting score lies in $[0,1]$, where higher values indicate greater semantic diversity. Specifically:

\moveup
\begin{equation*}
\begin{aligned}
\mathcal{D}_{\text{sem}}(\mathcal{A}^i_c)= \frac{1}{\binom{K}{2}} \sum_{1 \leq j < k \leq K}d_{\text{cos}}(\tilde{a}^i_{c,j}, \tilde{a}^i_{c,k}).
\end{aligned}
\end{equation*}
\moveup

where $\tilde{a}^i_{c,j}$ denotes the embedding of response $a^i_{c,j}$, and 
$d_{\text{cos}}()$ denotes normalized cosine distance.

\xhdr{Coverage Diversity}
While semantic diversity captures variation across responses, it may underestimate diversity in a consolidated answer that already covers multiple distinct perspectives for an open-ended question (see Appendix Table~\ref{tab:claim}). To address this gap, inspired by works on intra-answer diversity~\cite{zhang2025noveltybench,wright2025epistemic}, we propose a simplified \textit{coverage diversity} metric, which measures whether different responses cover distinct plausible answers in their content using the fraction of mutually non-overlapping aspects. 

Specifically, we use an LLM-based claim extractor $f$ to decompose each generated response into a set of \textit{atomic claims}, each representing a minimal, self-contained aspect that addresses the query. We then aggregate claims across the $K$ responses and compute the fraction of non-overlapping claims:

\moveup
\moveup
{\small
\begin{equation*}
\label{eq:viewpoint_div}
\begin{aligned}
\mathcal{C}^i_c &= \bigcup_{j=1}^{K} f(a^i_{c,j})
= \{c^i_{c,1}, \ldots, c^i_{c,k_i}\},
\mathcal{D}_{\text{Cov}}(\mathcal{A}^i_c)
= \frac{\left| \mathrm{uni}(\mathcal{C}^i_c) \right|}{k_i}
\end{aligned}
\end{equation*}
}
\moveup
\moveup

Here, $k_i$ is the total number of extracted claims for query $i$, and $\left| \mathrm{uni}(\cdot)\right|$ counts claims whose pairwise embedding similarity is below a predefined threshold $\tau$, treating them as distinct claims. Additional details are provided in Appendix~\ref{app: viewpoint}.

\subsection{Quality Metric}
\label{sec: quality}
\xhdr{Quality Score} Open-ended information-seeking queries may have an unbounded space of reasonable answers, making it impractical to pre-collect valid responses as ground truth. 
Thus, ground-truth-based metrics such as \textit{Factual Correctness} or \textit{Accuracy}~\cite{es2024ragas} are not applicable. Following prior work~\cite{badshah2024reference,yu2025improve,gu2024survey},
we adopt an \textit{LLM-as-a-Judge} framework for evaluation, which is better suited than reward models for uncertain settings without ground truth~\cite{xu2025ask}. We evaluate quality from four dimensions:
\emph{factual accuracy}, \emph{evidence support}, \emph{internal consistency}, \emph{question relevance}.
Judgments are reported on a five-level ordinal scale, with higher being better (\cf \S~\ref{app: quality}).

\xhdr{Human Annotation Agreement} 
To validate the LLM-as-a-Judge framework, we compare its scores against pre-collected human annotations (\cf \S~\ref{app:HAA}).
The results show that it achieves agreement comparable to human annotators, falling within the range of human judgment variability for this inherently subjective task, suggesting that our quality metric provides a reasonable proxy for human evaluation.

\subsection{Unified Metrics}
\label{sec: unified}

Diversity and quality naturally form a trade-off: broader answer coverage can introduce noise, while optimizing solely for quality may yield homogenized outputs.
Although both dimensions are important, assessing them separately makes it difficult to intuitively understand how well a model balances the diversity--quality trade-off. Inspired by prior work on harmonic aggregation for balancing competing objectives~\cite{sasaki2007truth,min2020ambigqa}, we introduce the \textit{Unified Diversity--Quality Harmonic Score} metric for capturing the diversity--quality trade-off. Specifically, it is defined as:

\moveup
\moveup
{\small
\begin{equation*}
\mathrm{U}_{\mathrm{Q}}^{\mathrm{D}}
= \frac{1}{N} \sum_{i=1}^{N} \mathcal{U}_{\mathrm{Q}}^{\mathrm{D}}\left(\mathcal{A}^i_c\right)
= \frac{1}{N} \sum_{i=1}^{N}
\frac{2 \cdot \tilde{Q}\left(\mathcal{A}^i_c\right) \cdot \tilde{D}\left(\mathcal{A}^i_c\right)}
{\tilde{Q}\left(\mathcal{A}^i_c\right) + \tilde{D}\left(\mathcal{A}^i_c\right)}.
\end{equation*}
}
\moveup
\moveup

where $\tilde{Q}^{i}$ and $\tilde{D}^{i}$ denote the query-wise min-max normalized quality and diversity scores in $[0,1]$. 
This normalization accounts for differences in the scales of quality and diversity scores, as well as query-level variation in difficulty and answer-space breadth. Using the two complementary diversity metrics from \S~\ref{sec: div}, semantic and coverage diversity, we obtain two complementary unified metrics, \textit{Unified Semantic Diversity–Quality Trade-off Score} $(\mathrm{U}^{\text{Sem}}_{\text{Q}})$ and \textit{Unified Coverage Diversity–Quality Trade-off Score} $(\mathrm{U}^{\text{Cov}}_{\text{Q}})$.

 \begin{table*}[t]
\centering
\tabcaption{
Overview of \textit{Unified Scores} under two complementary diversity measures: 
Semantic $\mathrm{U}^{\text{Sem}}_{\text{Q}}$ measures diversity of the overall response, while Coverage $\mathrm{U}^{\text{Cov}}_{\text{Q}}$ captures diversity over the plausible answer space covered by responses.
Best are shown in \textbf{bold}. 
\method achieves the \textbf{strongest} performance on \textit{both} metrics across \textit{all} models and datasets, indicating a consistently favorable diversity--quality trade-off. 
Full results are provided in \S~\ref{app:full_results}.
}
\vspace{1mm}
\small
\setlength{\tabcolsep}{3.2pt}
\renewcommand{\arraystretch}{1.3}
\resizebox{\textwidth}{!}{
\begin{tabular}{
l|
cc cc cc cc|
cc cc cc cc
}
\toprule
& \multicolumn{8}{c|}{\bf Infinity-Chat~\cite{jiang2025artificial}}
& \multicolumn{8}{c}{\bf IssueBench~\cite{rottger2025issuebench}} \\
\cmidrule(l{2pt}r{2pt}){2-9}
\cmidrule(l{2pt}r{2pt}){10-17}
\textbf{Methods}
& \multicolumn{2}{c}{\texttt{GPT-OSS-120B}}
& \multicolumn{2}{c}{\texttt{Qwen3-235B}}
& \multicolumn{2}{c}{\texttt{GPT-5-mini}}
& \multicolumn{2}{c|}{\texttt{GPT-5.1}}
& \multicolumn{2}{c}{\texttt{GPT-OSS-120B}}
& \multicolumn{2}{c}{\texttt{Qwen3-235B}}
& \multicolumn{2}{c}{\texttt{GPT-5-mini}}
& \multicolumn{2}{c}{\texttt{GPT-5.1}} \\
\cmidrule(lr){2-3}
\cmidrule(lr){4-5}
\cmidrule(lr){6-7}
\cmidrule(lr){8-9}
\cmidrule(lr){10-11}
\cmidrule(lr){12-13}
\cmidrule(lr){14-15}
\cmidrule(lr){16-17}
& Sem~$\uparrow$ & Cov~$\uparrow$
& Sem~$\uparrow$ & Cov~$\uparrow$
& Sem~$\uparrow$ & Cov~$\uparrow$
& Sem~$\uparrow$ & Cov~$\uparrow$
& Sem~$\uparrow$ & Cov~$\uparrow$
& Sem~$\uparrow$ & Cov~$\uparrow$
& Sem~$\uparrow$ & Cov~$\uparrow$
& Sem~$\uparrow$ & Cov~$\uparrow$
\\
\midrule

\textbf{Closed-Book LLMs} & & & & & & & & & & & & & & & & \\

Independent Sampling
& 0.362 & 0.362 & 0.249 & 0.216 & 0.119 & 0.417 & 0.094 & 0.346
& 0.360 & 0.487 & 0.250 & 0.170 & 0.164 & 0.345 & 0.116 & 0.322 \\

List Generation
& 0.262 & 0.267 & 0.247 & 0.236 & 0.167 & 0.160 & 0.456 & 0.518
& 0.509 & 0.526 & 0.359 & 0.373 & 0.369 & 0.392 & 0.583 & 0.644 \\

Iterative Generation
& 0.165 & 0.368 & 0.469 & 0.551 & 0.324 & 0.556 & 0.198 & 0.450
& 0.459 & 0.448 & 0.490 & 0.659 & 0.214 & 0.383 & 0.290 & 0.528 \\

\makecell[l]{Verbalized Sampling\\\cite{zhang2025verbalized}}
& 0.386 & 0.364 & 0.430 & 0.410 & 0.292 & 0.273 & 0.425 & 0.586
& 0.525 & 0.590 & 0.360 & 0.372 & 0.450 & 0.465 & 0.615 & 0.676 \\

\noalign{\vskip 0.5ex}\cdashline{1-17}\noalign{\vskip 0.75ex}

\textbf{RAGs} & & & & & & & & & & & & & & & & \\

Vanilla RAG
& 0.079 & 0.224 & 0.085 & 0.208 & 0.132 & 0.319 & 0.148 & 0.329
& 0.171 & 0.280 & 0.087 & 0.232 & 0.115 & 0.219 & 0.116 & 0.243 \\

\hspace{0.4em} + Diverse Re-ranking
& 0.105 & 0.240 & 0.059 & 0.216 & 0.145 & 0.330 & 0.151 & 0.334
& 0.190 & 0.307 & 0.085 & 0.254 & 0.158 & 0.284 & 0.172 & 0.332 \\

\hspace{0.4em} + Contexts Shuffle
& 0.137 & 0.281 & 0.158 & 0.286 & 0.172 & 0.361 & 0.179 & 0.340
& 0.260 & 0.375 & 0.160 & 0.315 & 0.196 & 0.341 & 0.221 & 0.348 \\

\hspace{0.4em} + Multi-Query
& 0.055 & 0.211 & 0.069 & 0.138 & 0.124 & 0.283 & 0.137 & 0.292
& 0.180 & 0.283 & 0.076 & 0.220 & 0.146 & 0.324 & 0.137 & 0.305 \\

\hspace{0.4em} + All
& 0.125 & 0.287 & 0.158 & 0.225 & 0.159 & 0.322 & 0.169 & 0.353
& 0.272 & 0.406 & 0.152 & 0.318 & 0.216 & 0.339 & 0.216 & 0.381 \\

\noalign{\vskip 0.5ex}\cdashline{1-17}\noalign{\vskip 0.75ex}

\textbf{\method}
& \textbf{0.434} & \textbf{0.706} & \textbf{0.488} & \textbf{0.669} & \textbf{0.557} & \textbf{0.728} & \textbf{0.473} & \textbf{0.713}
& \textbf{0.673} & \textbf{0.795} & \textbf{0.591} & \textbf{0.703} & \textbf{0.663} & \textbf{0.719} & \textbf{0.651} & \textbf{0.765} \\

\bottomrule
\end{tabular}
}
\moveup
\moveup
\label{tab:unified_summary}
\end{table*}

\begin{figure*}
    \centering
    \vspace{0.2cm}
    \moveup
    \includegraphics[width=1\linewidth]{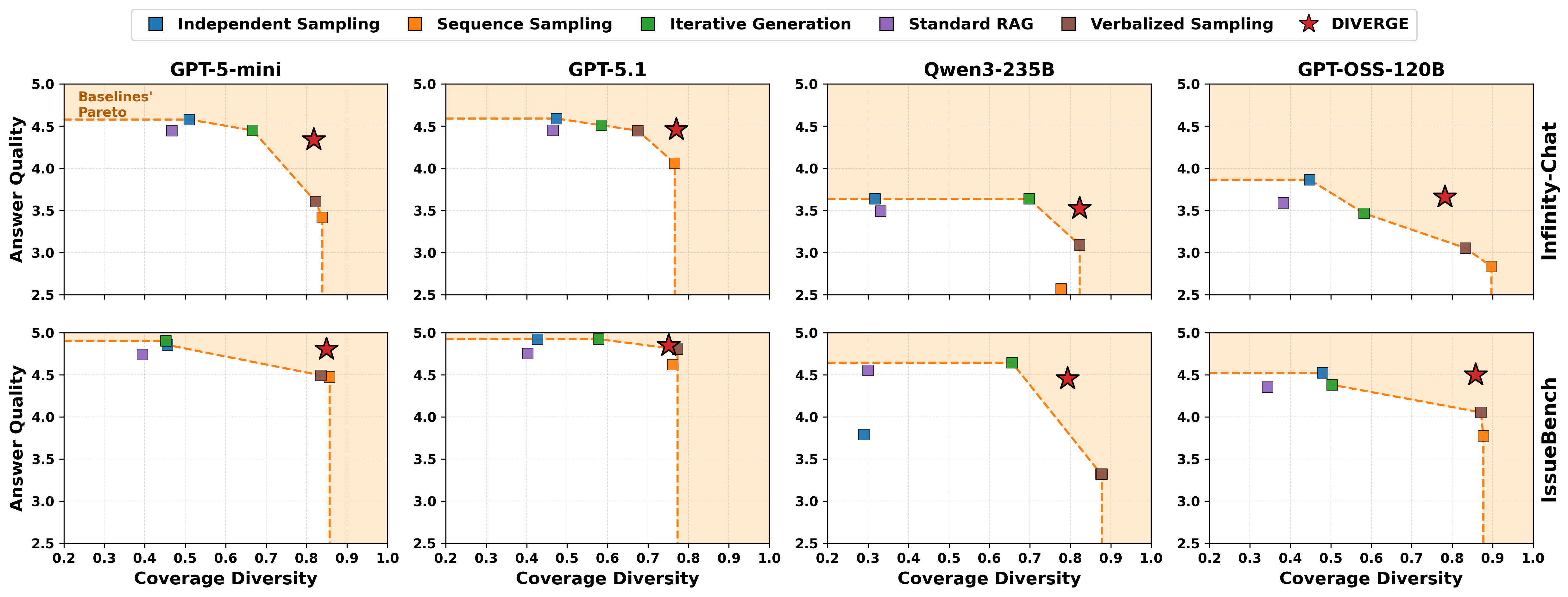}
    \moveup
    \moveup
    \moveups
    \figcaption{
        Visualization of the coverage diversity--quality trade-off across methods. 
        \textcolor{plotorange}{Orange lines} indicate the Pareto frontiers computed over the \textit{baseline methods} as
        reference. The \emph{Upper-right} region indicates a better trade-off. 
        Existing baselines either yield limited diversity or improve diversity at the expense of quality.
        Compared with \textit{direct independent prompting}, \method improves diversity by \textbf{$\sim2\times$} with only negligible quality degradation, making it the \textbf{only} method that consistently exhibits a favorable diversity--quality trade-off across models and datasets.
    }
    \label{fig: res}
    \moveups
\end{figure*}

\section{Experiment}
We first describe the datasets, baselines, and experimental setup, then present the results.

\subsection{Experimental Setup}
\label{sec: baselines}
\label{sec: dataset}

\xhdr{Datasets} We use two real-world open-ended QA datasets: \emph{Infinity-Chat}~\cite{jiang2025artificial} and \emph{IssueBench}~\cite{rottger2025issuebench}\footnote{Links: \href{https://huggingface.co/datasets/liweijiang/infinite-chats-taxonomy}{Infinity-Chat}; \href{https://huggingface.co/datasets/Paul/IssueBench}{IssueBench}; Details: ~\S~\ref{app:datasets}}. For each dataset, we sample $N=100$ queries and generate $K=10$ responses per query, yielding $1,000$ responses per model per dataset. \emph{Infinity-Chat} serves as a meta-benchmark of real-world open-ended prompts derived from \emph{WildChat}~\cite{zhao2024wildchat}, spanning \emph{10} diverse question categories (e.g., \textit{Decision Support}, \textit{Controversial Questions}, \textit{Skill Development}), covering a broad range of information-seeking scenarios. \emph{IssueBench} consists of open-ended questions from political and social domains, where diverse viewpoints are most critical, and the risk of homogenization is highest. We exclude \emph{NoveltyBench}~\cite{zhang2025noveltybench} and \emph{CoverageQA}~\cite{wong2024simplestrat}, as they are relatively simple and can often be answered without retrieval, making them unsuitable for evaluating our task.

\xhdr{Baselines}
Our baselines include two groups: (1) \textit{Closed-Book LLMs}, covering LLMs without retrieval and prompt-based diversity strategies~\cite{zhang2025verbalized}; and (2) \textit{Retrieval-Augmented Baselines}, covering vanilla RAG and common retrieval-side diversity enhancements. Detailed descriptions are provided in Appendix~\ref{app: baselines}.

For \textit{Closed-Book LLMs} baselines, we use (1) \underline{Independent Sampling}, the standard and most widely used setup in which the LLM is run $ K$ times independently to generate one response per run. We further consider three prompt-based diversity strategies: (2) \underline{List Generation}, which prompts the LLM once to return $K$ distinct bull-points; (3) \underline{Iterative Generation}, which appends previous responses to the dialogue history and asks for a new answer at each step; and (4) \underline{Verbalized Sampling}~\cite{zhang2025verbalized}, a SoTA strategy that prompts the model to generate $K$ candidates with verbalized probabilities, encouraging less typical outputs and reducing collapse.

For \textit{Retrieval-Augmented Baselines}, since \emph{no} existing RAG methods are designed for diverse open-ended information-seeking tasks, we adapt common retrieval-side diversity techniques to construct strong baselines. We include \underline{Vanilla RAG} and four variants: (1) \underline{Diversity Reranking}, which applies MMR~\cite{carbonell1998use}; (2) \underline{Context Shuffle}, which randomly shuffles contexts to reduce positional bias~\cite{liu2023lost}; (3) \underline{Multi-Query}, which retrieves using multiple LLM-generated query rewrites; and (4) \underline{All}, which combines above strategies. We also report additional baselines in Appendix~\ref{app: add_baselines}, showing that naively combining prompt-based diversity strategies with RAG yields unsatisfactory performance.

\xhdr{Backbone Models}
We evaluate four frontier reasoning-capable LLM backbones across different scales: two open-source models, \texttt{Qwen3-235B}~\cite{yang2025qwen3} and \texttt{GPT-OSS-120B}~\cite{agarwal2025gpt}, and two closed-source models, \texttt{GPT-5-mini} and \texttt{GPT-5.1}~\cite{openai2025gpt5systemcard}. 
This selection reflects realistic deployment settings under varying budget constraints. Details are provided in \S~\ref{app: setup}.

\subsection{Main Results}
\label{sec: main_res}
Table~\ref{tab:unified_summary} summarizes the \textit{Unified scores} comparison, and Figure~\ref{fig: res} visualizes the diversity--quality trade-off. Detailed results with full diversity and quality results are provided in Appx.~\ref{app: supp}. Based on these results, our findings can be summarized as follows:

\begin{itemize}[leftmargin=*, itemsep=0pt, topsep=0pt]
  \item \textbf{Standard RAG and retrieval-side diversity techniques do not yield more diverse outputs}: Compared to \textit{Independent Sampling LLMs}, RAGs even reduce diversity. This trend persists under simple retrieval diversity strategies (\S~\ref{sec: baselines}). As a result, these baselines fail to improve the overall \textit{Unified Scores}.
 \item \textbf{Prompt-based strategies increase diversity but often hurt quality}: \textit{List Generation} and \textit{Verbalized Sampling} improve diversity but substantially degrade quality, especially for weaker models. \textit{Iterative Generation} better preserves quality but yields only limited gains in diversity. Overall, these methods bring limited improvements.
\item \textbf{\method achieves the best trade-off}: It obtains the highest \textit{Unified Score} across all datasets and models. Compared to \textit{Independent Sampling}, \method improves both semantic diversity and coverage diversity by around \textbf{2$\times$}, while maintaining comparable quality with only a marginal score drop ($<2\%$) on average.
\end{itemize}

\section{Further Analysis}
\label{sec: ana}

 %Additional metric validation, including claim extraction reliability, potential judge bias, and score stability, is provided in Appendix~\ref{app:metric_validation}.
 
We further analyze \method from four perspectives: component ablations, threshold sensitivity, case study, and metric correlations. Due to space constraints, additional analyses are provided in Appendix~\ref{app:add_ana}, including extended metric validation, full case-study results, latency and token-cost analysis, and robustness analyses across different settings.

\begin{figure}[t!]
    \centering
    \moveups
    \includegraphics[
        width=\linewidth,
    ]{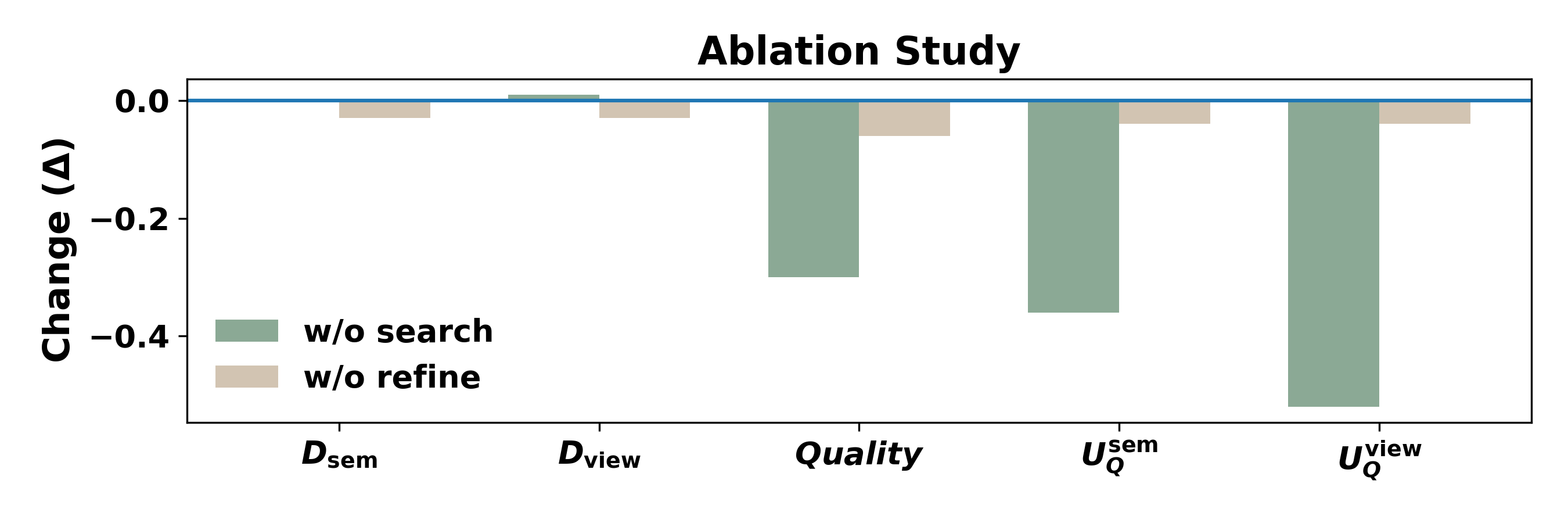}
    \moveup
    \moveup
    \moveup
    \figcaption{Ablation study of \method showing the impact of removing search grounding and result refinement on the \textit{GPT-5-mini} model, highlighting the contributions of these components in \method.}
    \label{fig:ablation}
    \moveups
\end{figure}

\xhdr{Ablation Study} 
We analyze the effects of removing key components of \method, including search grounding and result refinement. As shown in Figure~\ref{fig:ablation}, removing either component leads to a noticeable degradation in \emph{Quality}, which in turn results in a lower \emph{Unified Score}. These results empirically demonstrate the effectiveness of both components in achieving a favorable diversity–quality trade-off. 

\xhdr{Threshold Sensitivity Analysis}
We examine the sensitivity of the coverage diversity threshold $\tau$ over $\{0.70, 0.75, 0.80, 0.85\}$, covering a moderate-to-high range explored in prior embedding-similarity threshold analyses and applications~\cite{rekabsaz2017exploration,gohsen2023paraphrase}.
Across all settings, \method consistently outperforms all baselines in \emph{Unified Score}, and the ranking on coverage diversity of methods remains highly stable across thresholds, with Kendall's $W=0.95$, suggesting that our conclusions are not sensitive to the choice of $\tau$.

\xhdr{Case Study}
We further provide qualitative error analyses and PCA visualizations to illustrate the strengths and remaining challenges of \method. 

The PCA visualization in Appendix Figure~\ref{fig:case_study} further illustrates this trend: responses from direct LLM prompting and RAGs are concentrated in narrow regions of the projected embedding space, while \method exhibits a wider spread, indicating broader exploration of diverse solution directions.

We also analyze 30 low-quality outputs from \method. As shown in Appendix Table~\ref{tab:error_cases}, these cases mainly involve partial deviation from user intent (40\%), overly generic recommendations (30\%), and overemphasis on peripheral aspects (17\%), which provide useful guidance for future extensions of \method, particularly toward more precise viewpoint selection and integration.

\xhdr{Metric Correlation Analysis}
Analysis in Appendix~\ref{app:correlation_analysis} further supports our metric design. As shown in Figures~\ref{fig:correlation} and~\ref{fig: relation}, all \textit{diversity} metrics are negatively correlated with \textit{quality}, while semantic diversity and coverage diversity are positively correlated but capture complementary signals.  In particular, aggregated responses containing more distinct internal perspectives tend to exhibit higher coverage diversity relative to semantic diversity, which is consistent with our hypothesis that coverage diversity better reflects intra-response diversity.

\vspace{-1em}
\section{Conclusion}
We identify a key limitation of existing RAG systems: they often underutilize diverse retrieved contexts and produce homogenized outputs for open-ended information seeking. We propose \method, a plug-and-play agentic RAG framework that grounds retrieval and generation in diverse viewpoints. Experiments show that \method improves diversity while preserving quality, highlighting the need to optimize the diversity--quality trade-off in open-ended information-seeking tasks.

\clearpage

\section*{Limitations}
Evaluating open-ended responses remains challenging, especially for knowledge-seeking tasks where even human experts may disagree. Although we use LLM-as-a-judge evaluation, quality assessment may still be imperfect, and it remains unclear which diversity metrics best align with human perception. Future work should incorporate broader human evaluation and develop domain-specific benchmarks with multiple plausible answers. 

In an ideal setting, one would conduct large-scale human evaluations to measure the consistency between diversity metrics, overall quality trade-offs, and human judgments. However, evaluating diversity in open-ended tasks is inherently subjective and highly challenging even for human annotators. As a result, obtaining reliable judgments would require more participants, careful user study design, and substantially higher annotation costs, which were beyond our current budget constraints. We hope future work can further explore this direction through broader human evaluation.

Although our method improves result diversity and supports more inclusive and diverse information seeking, it may still underrepresent minority perspectives and communities. Moreover, in high-stakes domains such as medical or financial advice, seeking diverse information may increase the risk of exposing users to information that is not fully accurate. Therefore, applying our method to such domains may require practitioners to adopt domain-specific risk checks beyond the quality metrics considered in this paper.

\section*{Acknowledgements}

We thank Yiju Guo, Xinhao Nie, and Nearchos Potamitis for
insightful discussions and for reviewing an initial draft of
this paper. Arora’s lab is partly supported by grants from the
Novo Nordisk Foundation (NNF24OC0099109), the Pioneer Centre for AI, and EU Horizon 2020 (101168951). We also gratefully acknowledge generous gifts from Microsoft and IT-vest - networking universities.

\bibliography{custom}
\appendix

\clearpage

\begin{algorithm*}[t]
\caption{\method}
\label{alg:divrag}
\begin{algorithmic}[1]
\REQUIRE Generator $LM$, Retriever $\mathcal{R}$, Web Agent $\mathcal{W}$, query $q$, generation size $K$
\ENSURE Response set $\mathcal{A} = \{a_1, \ldots, a_K\}$

\STATE Initialize memory $\mathcal{M} \gets \emptyset$
\STATE Initialize response set $\mathcal{A} \gets \emptyset$
\STATE Initialize viewpoint set $\mathcal{V} \gets \emptyset$

\vspace{0.2em}
\STATE \textbf{// Initial RAG generation}
\STATE $q_1 \gets q$
\STATE $\mathcal{I}_1 \gets \mathcal{W}(q_1)$ \hfill \COMMENT{Open-web search}
\STATE $d_1 \gets \mathcal{R}(\mathcal{I}_1)$ \hfill \COMMENT{Retrieve supporting evidence}
\STATE $\hat{a}_1 \gets LM_{\mathrm{gen}}(q, d_1)$ \hfill \COMMENT{Initial RAG response}
\STATE $a_1 \gets LM_{\mathrm{refine}}(q, \hat{a}_1)$ \hfill \COMMENT{Refine for quality}
\STATE $\mathcal{V}_1 \gets LM_{\mathrm{summarize}}(q, a_1)$ \hfill \COMMENT{Extract existing viewpoints}
\STATE $\mathcal{V} \gets \mathcal{V} \cup \mathcal{V}_1$
\STATE $\mathcal{M} \gets \mathcal{M} \cup \{(q_1, d_1, \mathcal{V}_1, a_1)\}$
\STATE $\mathcal{A} \gets \mathcal{A} \cup \{a_1\}$

\vspace{0.2em}
\STATE \textbf{// Iterative diversity-oriented generation}
\FOR{$t = 2$ to $K$}
    \STATE $v_t \gets LM_{\mathrm{reflect}}(q, \mathcal{V}, \mathcal{M})$
    \hfill \COMMENT{Identify an uncovered viewpoint}
    
    \STATE $q_t \gets LM_{\mathrm{query}}(q, v_t)$
    \hfill \COMMENT{Generate a viewpoint-conditioned query}
    
    \STATE $\mathcal{I}_t \gets \mathcal{W}(q_t)$
    \hfill \COMMENT{Search the open web}
    
    \STATE $d_t \gets \mathcal{R}(\mathcal{I}_t, q_t, \mathcal{M})$
    \hfill \COMMENT{Diversity-aware retrieval and reranking}
    
    \STATE $\hat{a}_t \gets LM_{\mathrm{gen}}(q, v_t, d_t)$
    \hfill \COMMENT{Viewpoint-conditioned generation}
    
    \STATE $a_t \gets LM_{\mathrm{refine}}(q, v_t, \hat{a}_t)$
    \hfill \COMMENT{Refine for relevance and coherence}
    
    \STATE $\mathcal{V} \gets \mathcal{V} \cup \{v_t\}$
    \STATE $\mathcal{M} \gets \mathcal{M} \cup \{(q_t, d_t, v_t, a_t)\}$
    \STATE $\mathcal{A} \gets \mathcal{A} \cup \{a_t\}$
\ENDFOR

\STATE \textbf{return} $\mathcal{A}$
\end{algorithmic}
\end{algorithm*}

\begin{figure*}[ht!]
    \centering
    \includegraphics[width=1\linewidth]{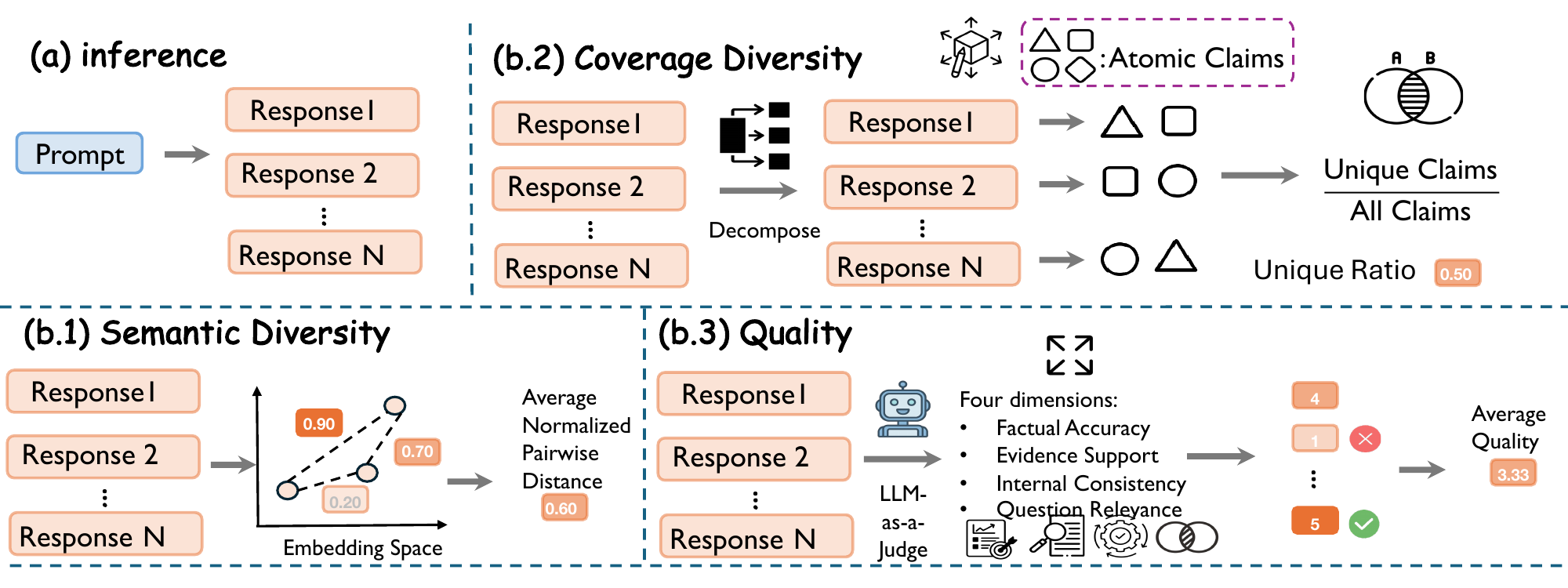}
    \caption{Illustrative examples of the diversity and quality metrics}
    \label{fig:metric_example}
\end{figure*}

\section{Additional Information of \method}
\label{app: method}

\subsection{Algorithm}
The algorithm block is stated in the Algorithm~\ref{alg:divrag}

\begin{table*}[t]
\centering

\resizebox{0.9\textwidth}{!}{
    \begin{tabular}{
    l
    cc c >{\columncolor{udqbg}}c >{\columncolor{udqbg}}c
    cc c >{\columncolor{udqbg}}c >{\columncolor{udqbg}}c
    }
    \toprule
    & \multicolumn{5}{c}{\bf GPT-5-mini}
    & \multicolumn{5}{c}{\bf GPT-5.1} 
    \\
    \cmidrule(l{2pt}r{2pt}){2-6}
    \cmidrule(l{2pt}r{2pt}){7-11}
    \textbf{Dataset: \texttt{Infinity-Chat}}
    & \multicolumn{2}{c}{\textbf{Diversity}~$\uparrow$}
    & \textbf{Quality}~$\uparrow$
    & \multicolumn{2}{>{\columncolor{udqbg}}c}{\textbf{Unified Score}~$\uparrow$}
    & \multicolumn{2}{c}{\textbf{Diversity}~$\uparrow$}
    & \textbf{Quality}~$\uparrow$
    & \multicolumn{2}{>{\columncolor{udqbg}}c}{\textbf{Unified Score
    ~$\uparrow$}}\\
    
    & $\mathcal{D}_{\text{Sem}}$ & $\mathcal{D}_{\text{Cov}}$ & & $\mathrm{U}^{\text{Sem}}_{\text{Q}}$ & $\mathrm{U}^{\text{Cov}}_{\text{Q}}$ & $\mathcal{D}_{\text{Sem}}$ & $\mathcal{D}_{\text{Cov}}$ & & $\mathrm{U}^{\text{Sem}}_{\text{Q}}$ & $\mathrm{U}^{\text{Cov}}_{\text{Q}}$  \\
    
    \midrule
    \textbf{Closed-Book LLMs} & & & & & & & & & & \\
    
    Independent Sampling
    & 0.100 & 0.510 & 4.578 & 0.119 & 0.417 & 0.096 & 0.474 & 4.590 & 0.094 & 0.346 \\
    List Generation & \cellcolor{red!20}0.446 & \cellcolor{red!20}0.839 & \cellcolor{neg!20}3.417 & 0.167 & 0.160 & \cellcolor{red!20}0.309 & \cellcolor{red!20}0.766 & \cellcolor{neg!20}4.059 & 0.456 & 0.518  \\
    Iterative Generation & \cellcolor{red!5}0.176 & \cellcolor{red!10}0.667 & 4.449 & 0.324 & 0.556 & \cellcolor{red!5}0.131 & 0.585 & 4.510 & 0.198 & 0.450 \\
    Verbalized Sampling~\cite{zhang2025verbalized} & \cellcolor{red!20}0.417 & \cellcolor{red!20}0.822 & \cellcolor{neg!20}3.603 & 0.292 & 0.273 & \cellcolor{red!20}0.217 & \cellcolor{red!20}0.675 & 4.447 & 0.425 & 0.586 \\
    
    \noalign{\vskip 0.5ex}\cdashline{1-11}\noalign{\vskip 0.75ex}
    
    \textbf{RAGs} & & & & & & & & & & \\

    Vanilla RAG & 0.106 & 0.467 & 4.444 & 0.132 & 0.319 & 0.107 & 0.465 & 4.449 & 0.148 & 0.329 \\
    \hspace{0.4em} + Diverse Re-ranking & 0.106 & 0.469 & 4.465 & 0.145 & 0.330 & 0.109 & 0.475 & 4.428 & 0.151 & 0.334 \\
    \hspace{0.4em} + Contexts Shuffle & 0.116 & 0.487 & 4.429 & 0.172 & 0.361 & 0.119 & 0.475 & 4.422 & 0.179 & 0.340 \\
    \hspace{0.4em} + Multi-Query & 0.100 & 0.446 & 4.423 & 0.124 & 0.283 & 0.102 & 0.445 & 4.452 & 0.137 & 0.292 \\
    \hspace{0.4em} + All & 0.110 & 0.466 & 4.464 & 0.159 & 0.322 & 0.110 & 0.471 & 4.532 & 0.169 & 0.353 \\

    \noalign{\vskip 0.5ex}\cdashline{1-11}\noalign{\vskip 0.75ex}
    \textbf{\method}  
    & \cellcolor{red!15}0.269 & \cellcolor{red!20}0.818 & 4.342 & \textbf{0.557} & \textbf{0.728}
    & \cellcolor{red!20}0.219 & \cellcolor{red!20}0.770 & 4.462 & \textbf{0.473} & \textbf{0.713}\\
    \bottomrule
    \end{tabular}
}
\tabcaption{Evaluation of diversity and quality for \textbf{Close-sourced LLMs} on the \texttt{Infinity-Chat}. 
(1) The diversity-quality trade-off is presented using \textit{Independent} generation as the baseline. \good{Red} indicates statistically significant improvements and \bad{blue} indicates statistically significant degradations ($p<0.01$), with darker colors indicating larger effect sizes. \method is the \emph{only} method that improves diversity while maintaining comparable quality.
(2) \emph{Unified Score}, highlighted in {\color{udqbg} purple}, provides an overall comparison by jointly accounting for diversity and quality, best shown in \textbf{bold}.
\method achieves the strongest performance across all models.} 
\label{tab:full_infinity_closed}
\end{table*}

\begin{table*}
\centering

\resizebox{0.9\textwidth}{!}{
    \begin{tabular}{
    l
    cc c >{\columncolor{udqbg}}c >{\columncolor{udqbg}}c
    cc c >{\columncolor{udqbg}}c >{\columncolor{udqbg}}c
    }
    \toprule
    & \multicolumn{5}{c}{\bf GPT-OSS-120B}
    & \multicolumn{5}{c}{\bf Qwen3-235B} 
    \\
    \cmidrule(l{2pt}r{2pt}){2-6}
    \cmidrule(l{2pt}r{2pt}){7-11}
    \textbf{Dataset: \texttt{Infinity-Chat}}
    & \multicolumn{2}{c}{\textbf{Diversity}~$\uparrow$}
    & \textbf{Quality}~$\uparrow$
    & \multicolumn{2}{>{\columncolor{udqbg}}c}{\textbf{Unified Score}~$\uparrow$}
    & \multicolumn{2}{c}{\textbf{Diversity}~$\uparrow$}
    & \textbf{Quality}~$\uparrow$
    & \multicolumn{2}{>{\columncolor{udqbg}}c}{\textbf{Unified Score}~$\uparrow$}\\
    
    & $\mathcal{D}_{\text{Sem}}$ & $\mathcal{D}_{\text{Cov}}$ & 
    & $\mathrm{U}^{\text{Sem}}_{\text{Q}}$ & $\mathrm{U}^{\text{Cov}}_{\text{Q}}$
    & $\mathcal{D}_{\text{Sem}}$ & $\mathcal{D}_{\text{Cov}}$ & 
    & $\mathrm{U}^{\text{Sem}}_{\text{Q}}$ & $\mathrm{U}^{\text{Cov}}_{\text{Q}}$  \\
    
    \midrule
    \textbf{Closed-Book LLMs} & & & & & & & & & & \\
    
    Independent Sampling
    & 0.139 & 0.448 & 3.863 & 0.362 & 0.362
    & 0.136 & 0.317 & 3.637 & 0.249 & 0.216 \\
    
    List Generation
    & \cellcolor{red!20}0.515 & \cellcolor{red!20}0.897 & \cellcolor{neg!20}2.834 & 0.262 & 0.267
    & \cellcolor{red!20}0.410 & \cellcolor{red!20}0.777 & \cellcolor{neg!20}2.569 & 0.247 & 0.236 \\
    
    Iterative Generation
    & \cellcolor{red!10}0.196 & \cellcolor{red!5}0.582 & \cellcolor{neg!10}3.465 & 0.165 & 0.368
    & \cellcolor{red!10}0.207 & \cellcolor{red!10}0.698 & 3.636 & 0.469 & 0.551 \\
    
    Verbalized Sampling~\cite{zhang2025verbalized}
    & \cellcolor{red!20}0.488 & \cellcolor{red!20}0.833 & \cellcolor{neg!20}3.053 & 0.386 & 0.364
    & \cellcolor{red!20}0.461 & \cellcolor{red!20}0.823 & \cellcolor{neg!20}3.090 & 0.430 & 0.410 \\
    
    \noalign{\vskip 0.5ex}\cdashline{1-11}\noalign{\vskip 0.75ex}
    
    \textbf{RAGs} & & & & & & & & & & \\

    Vanilla RAG
    & 0.089 & \cellcolor{neg!10}0.383 & 3.589 & 0.079 & 0.224
    & 0.078 & 0.331 & 3.490 & 0.085 & 0.208 \\

    \hspace{0.4em} + Diverse Re-ranking
    & 0.103 & \cellcolor{neg!10}0.389 & 3.575 & 0.105 & 0.240
    & 0.068 & 0.315 & 3.524 & 0.059 & 0.216 \\

    \hspace{0.4em} + Contexts Shuffle
    & 0.109 & 0.420 & 3.562 & 0.137 & 0.281
    & 0.106 & 0.386 & 3.530 & 0.158 & 0.286 \\

    \hspace{0.4em} + Multi-Query
    & 0.083 & \cellcolor{neg!10}0.360 & 3.766 & 0.055 & 0.211
    & 0.063 & 0.297 & 3.486 & 0.069 & 0.138 \\

    \hspace{0.4em} + All
    & 0.109 & 0.452 & 3.698 & 0.125 & 0.287
    & 0.091 & 0.367 & 3.597 & 0.158 & 0.225 \\

    \noalign{\vskip 0.5ex}\cdashline{1-11}\noalign{\vskip 0.75ex}
    
    \textbf{\method}
    & \cellcolor{red!20}0.230 &  \cellcolor{red!20}0.782 & 3.660 & \textbf{0.434} & \textbf{0.706}
    & \cellcolor{red!20} 0.259 & \cellcolor{red!20} 0.823 & 3.524 & \textbf{0.488} & \textbf{0.669} \\
    
    \bottomrule
    \end{tabular}
}
\tabcaption{Evaluation of diversity and quality for \textbf{Open-sourced LLMs} on the \texttt{Infinity-Chat}. Details of the formatting and notation information can be found in Table~\ref{tab:full_infinity_closed}.} 
\label{tab:full_infinity_open}
\moveup
\end{table*}

\begin{table*}[t]
\centering

\resizebox{0.9\textwidth}{!}{
    \begin{tabular}{
    l
    cc c >{\columncolor{udqbg}}c >{\columncolor{udqbg}}c
    cc c >{\columncolor{udqbg}}c >{\columncolor{udqbg}}c
    }
    \toprule
    & \multicolumn{5}{c}{\bf GPT-5-mini}
    & \multicolumn{5}{c}{\bf GPT-5.1} 
    \\
    \cmidrule(l{2pt}r{2pt}){2-6}
    \cmidrule(l{2pt}r{2pt}){7-11}
    \textbf{Dataset: \texttt{IssueBench}}
    & \multicolumn{2}{c}{\textbf{Diversity}~$\uparrow$}
    & \textbf{Quality}~$\uparrow$
    & \multicolumn{2}{>{\columncolor{udqbg}}c}{\textbf{Unified Score}~$\uparrow$}
    & \multicolumn{2}{c}{\textbf{Diversity}~$\uparrow$}
    & \textbf{Quality}~$\uparrow$
    & \multicolumn{2}{>{\columncolor{udqbg}}c}{\textbf{Unified Score}~$\uparrow$}\\
    
    & $\mathcal{D}_{\text{Sem}}$ & $\mathcal{D}_{\text{Cov}}$ & 
    & $\mathrm{U}^{\text{Sem}}_{\text{Q}}$ & $\mathrm{U}^{\text{Cov}}_{\text{Q}}$
    & $\mathcal{D}_{\text{Sem}}$ & $\mathcal{D}_{\text{Cov}}$ & 
    & $\mathrm{U}^{\text{Sem}}_{\text{Q}}$ & $\mathrm{U}^{\text{Cov}}_{\text{Q}}$  \\
    
    \midrule
    \textbf{Closed-Book LLMs} & & & & & & & & & & \\
    
    Independent Sampling
    & 0.137 & 0.456 & 4.853 & 0.164 & 0.345
    & 0.124 & 0.427 & 4.920 & 0.116 & 0.322 \\
    
    List Generation
    & \cellcolor{red!20}0.400 & \cellcolor{red!20}0.857 & \cellcolor{neg!10}4.472 & 0.369 & 0.392
    & \cellcolor{red!20}0.266 & \cellcolor{red!20}0.761 & \cellcolor{neg!10}4.618 & 0.583 & 0.644 \\
    
    Iterative Generation
    & 0.147 & 0.452 & 4.902 & 0.214 & 0.383
    & 0.164 & 0.578 & 4.922 & 0.290 & 0.528 \\
    
    Verbalized Sampling~\cite{zhang2025verbalized}
    & \cellcolor{red!20}0.387 & \cellcolor{red!20}0.836 & \cellcolor{neg!10}4.492 & 0.450 & 0.465
    & \cellcolor{red!20}0.312 & \cellcolor{red!20}0.773 & 4.802 & 0.615 & 0.676 \\
    
    \noalign{\vskip 0.5ex}\cdashline{1-11}\noalign{\vskip 0.75ex}
    
    \textbf{RAGs} & & & & & & & & & & \\

    Vanilla RAG
    & 0.126 & 0.394 & 4.738 & 0.115 & 0.219
    & 0.126 & 0.403 & 4.752 & 0.116 & 0.243 \\

    \hspace{0.4em} + Diverse Re-ranking
    & 0.130 & 0.418 & 4.714 & 0.158 & 0.284
    & 0.130 & 0.429 & 4.742 & 0.172 & 0.332 \\

    \hspace{0.4em} + Contexts Shuffle
    & 0.151 & 0.433 & 4.748 & 0.196 & 0.341
    & 0.152 & 0.420 & 4.770 & 0.221 & 0.348 \\

    \hspace{0.4em} + Multi-Query
    & 0.124 & 0.415 & 4.824 & 0.146 & 0.324
    & 0.122 & 0.395 & 4.822 & 0.137 & 0.305 \\

    \hspace{0.4em} + All
    & 0.152 & 0.461 & 4.911 & 0.216 & 0.339
    & 0.141 & 0.450 & 4.790 & 0.216 & 0.381 \\

    \noalign{\vskip 0.5ex}\cdashline{1-11}\noalign{\vskip 0.75ex}
    
    \textbf{\method}  
    & \cellcolor{red!20}0.349 & \cellcolor{red!20}0.849 & 4.803 & \textbf{0.663} & \textbf{0.719}
    & \cellcolor{red!20}0.293 & \cellcolor{red!20}0.751 & 4.848 & \textbf{0.651} & \textbf{0.765}\\
    
    \bottomrule
    \end{tabular}
}
\tabcaption{Evaluation of diversity and quality for \textbf{Close-sourced LLMs} on the \texttt{IssueBench}. Details of the formatting and notation information can be found in Table~\ref{tab:full_infinity_closed}.} 
\label{tab:full_issue_closed}
\end{table*}

\begin{table*}
\centering
\resizebox{0.9\textwidth}{!}{
    \begin{tabular}{
    l
    cc c >{\columncolor{udqbg}}c >{\columncolor{udqbg}}c
    cc c >{\columncolor{udqbg}}c >{\columncolor{udqbg}}c
    }
    \toprule
    & \multicolumn{5}{c}{\bf GPT-OSS-120B}
    & \multicolumn{5}{c}{\bf Qwen3-235B-A22B} 
    \\
    \cmidrule(l{2pt}r{2pt}){2-6}
    \cmidrule(l{2pt}r{2pt}){7-11}
    \textbf{Dataset: \texttt{IssueBench}}
    & \multicolumn{2}{c}{\textbf{Diversity}~$\uparrow$}
    & \textbf{Quality}~$\uparrow$
    & \multicolumn{2}{>{\columncolor{udqbg}}c}{\textbf{Unified Score}~$\uparrow$}
    & \multicolumn{2}{c}{\textbf{Diversity}~$\uparrow$}
    & \textbf{Quality}~$\uparrow$
    & \multicolumn{2}{>{\columncolor{udqbg}}c}{\textbf{Unified Score}~$\uparrow$}\\
    
    & $\mathcal{D}_{\text{Sem}}$ & $\mathcal{D}_{\text{Cov}}$ & 
    & $\mathrm{U}^{\text{Sem}}_{\text{Q}}$ & $\mathrm{U}^{\text{Cov}}_{\text{Q}}$
    & $\mathcal{D}_{\text{Sem}}$ & $\mathcal{D}_{\text{Cov}}$ & 
    & $\mathrm{U}^{\text{Sem}}_{\text{Q}}$ & $\mathrm{U}^{\text{Cov}}_{\text{Q}}$  \\
    
    \midrule
    \textbf{Closed-Book LLMs} & & & & & & & & & & \\
    
    Independent Sampling
    & 0.191 & 0.480 & 4.521 & 0.360 & 0.487
    & 0.172 & 0.290 & 4.290 & 0.250 & 0.170 \\
    
    List Generation
    & \cellcolor{red!20}0.470 & \cellcolor{red!20}0.877 & \cellcolor{neg!20}3.774 & 0.509 & 0.526
    & \cellcolor{red!20}0.476 & \cellcolor{red!20}0.876 & \cellcolor{neg!20}3.320 & 0.359 & 0.373 \\
    
    Iterative Generation
    & 0.222 & 0.503 & 4.380 & 0.459 & 0.448
    & \cellcolor{red!10}0.240 & \cellcolor{red!20}0.656 & 4.542 & 0.490 & 0.659 \\
    
    Verbalized Sampling~\cite{zhang2025verbalized}
    & \cellcolor{red!20}0.426 & \cellcolor{red!20}0.871 & \cellcolor{neg!20}4.053 & 0.525 & 0.590
    & \cellcolor{red!20}0.480 & \cellcolor{red!20}0.878 & \cellcolor{neg!20}3.319 & 0.360 & 0.372 \\
    
    \noalign{\vskip 0.5ex}\cdashline{1-11}\noalign{\vskip 0.75ex}
    
    \textbf{RAGs} & & & & & & & & & & \\

    Vanilla RAG
    & 0.118 & 0.344 & 4.353 & 0.171 & 0.280
    & 0.087 & 0.300 & 4.552 & 0.087 & 0.232 \\
    
    \hspace{0.4em} + Diverse Re-ranking
    & 0.115 & 0.347 & 4.373 & 0.190 & 0.307
    & 0.085 & 0.311 & 4.526 & 0.085 & 0.254 \\
    
    \hspace{0.4em} + Contexts Shuffle
    & 0.152 & 0.413 & 4.328 & 0.260 & 0.375
    & 0.111 & 0.347 & 4.578 & 0.160 & 0.315 \\
    
    \hspace{0.4em} + Multi-Query
    & 0.110 & 0.332 & 4.324 & 0.180 & 0.283
    & 0.080 & 0.290 & 4.634 & 0.076 & 0.220 \\
    
    \hspace{0.4em} + All
    & 0.147 & 0.442 & 4.352 & 0.272 & 0.406
    & 0.104 & 0.350 & 4.625 & 0.152 & 0.318 \\

    \noalign{\vskip 0.5ex}\cdashline{1-11}\noalign{\vskip 0.75ex}
    
    \textbf{\method}  
    & \cellcolor{red!20}0.364 & \cellcolor{red!20}0.858 & 4.500 & \textbf{0.673} & \textbf{0.795}
    & \cellcolor{red!20}0.335 & \cellcolor{red!20}0.793 & 4.454 & \textbf{0.591} & \textbf{0.703}\\
    
    \bottomrule
    \end{tabular}
}
\tabcaption{Evaluation of diversity and quality for \textbf{Open-sourced LLMs} on the \texttt{IssueBench}. Details of the formatting and notation information can be found in Table~\ref{tab:full_infinity_closed}.} 
\label{tab:full_issue_open}
\moveup
\end{table*}

\subsection{Prompts}

The prompts used by \method are shown in Figure~\ref{fig:prompt_summary}--\ref{fig:prompt_refine_without_view}.

\subsection{Overview of \method}

\method is an iterative retrieval-augmented generation framework designed to produce
\emph{diverse yet relevant} answers to open-ended questions by explicitly modeling
historical retrievals and generated viewpoints.

\paragraph{Initialization.}
Given an input query $q$, \method initializes a diversity memory that stores:
(i) previously issued queries,
(ii) generated answers,
(iii) extracted viewpoints, and
(iv) embeddings of retrieved documents.
The embedding model, chunking strategy, and large language model (LLM) are configured
globally and shared across iterations.

\paragraph{First Iteration ($t=0$).}
\method begins with a standard retrieval-augmented generation step:
\begin{enumerate}
    \item \textbf{Retrieval.} The input query $q$ is used to perform a web search.
    Retrieved documents are chunked, embedded, and indexed into a vector store
    (cached per query for efficiency).
    
    \item \textbf{Diversity-Aware Reranking.} Retrieved documents are reranked using
    a diversity-aware postprocessor. Since no retrieval history exists at $t=0$,
    ranking is primarily driven by relevance.
    
    \item \textbf{Generation.} The LLM generates an answer grounded in the retrieved documents
    using a standard RAG prompt.
    
    \item \textbf{View Summarization.} The generated answer is summarized into a set of
    high-level viewpoints, which serve as semantic anchors for subsequent iterations.
\end{enumerate}

\paragraph{Subsequent Iterations ($t > 0$).}
For each subsequent iteration, \method explicitly encourages novel perspectives:
\begin{enumerate}
    \item \textbf{View Generation.} A new viewpoint is generated by prompting the LLM with
    the original question and the set of previously explored viewpoints.
    
    \item \textbf{Query Reformulation.} A new query is synthesized conditioned on the newly
    generated viewpoint, steering retrieval toward under-explored semantic regions.
    
    \item \textbf{History-Aware Retrieval.} Documents are retrieved and reranked using the DivReranker, which balances:
    (i) relevance to the current query,
    (ii) diversity among documents selected within the current iteration, and
    (iii) dissimilarity to documents retrieved in earlier iterations.
    
    \item \textbf{View-Conditioned Generation.} The LLM generates an answer grounded in the retrieved documents and explicitly framed from the specified viewpoint.
    
    \item \textbf{Memory Update.} The new query, retrieved document embeddings,
    generated answer, and viewpoint are stored in memory.
\end{enumerate}

\paragraph{Termination.}
The process repeats until a predefined number of generations $K$ is reached.
\method outputs a set of answers that are grounded in external evidence,
diverse across semantic viewpoints, and non-redundant with respect to past retrievals.

\subsection{Overview of Search in \method}
We implement a lightweight and reproducible web search and document extraction pipeline
to support retrieval-augmented generation.

\paragraph{Query Processing.}
Given a textual query, the system retrieves web pages using a Google/DuckDuckGo-based
search interface executed via a subprocess. For each query, the search module
requests up to $2N$ candidate URLs to account for filtering and extraction failures,
where $N$ is the target number of retained documents. To enhance reproducibility, we will provide a frozen corpus snapshot of the retrieval results on the Open Web.

\paragraph{Domain and Format Filtering.}
To improve content quality and reduce noise, retrieved URLs are filtered by:
(i) excluding social media and multimedia platforms (e.g., Twitter, YouTube, Instagram),
(ii) removing PDF documents, and
(iii) ignoring domains matching a predefined blocklist.
Only standard HTML pages from non-blacklisted domains are processed further.

\paragraph{HTML Content Extraction.}
For each retained URL, the system downloads the corresponding web page and extracts
raw textual content using an HTML parser.
Script, style, and non-textual elements are removed prior to extraction.
The remaining visible text is normalized by line stripping and concatenation.

Pages that fail to download, return access errors (e.g., HTTP 403), or yield insufficient results content is discarded.

\paragraph{Length Filtering.}
Extracted documents are required to exceed a minimum character threshold to ensure
sufficient informational content.
Only documents satisfying this constraint are retained as retrieval candidates.

\paragraph{Rate Control and Robustness.}
To reduce the risk of request throttling and blocking, the pipeline enforces randomized
delays between requests and executes all search operations in a subprocess-safe manner.
Errors during search or extraction are logged and handled gracefully without interrupting
batch processing.

\paragraph{Batch Processing and Output.}
For large-scale experiments, queries can be processed in batch from an input file.
For each query, the system outputs a list of retrieved documents, including the source URL,
extracted text, and document length.
All results are stored in a structured JSON format with timestamps to ensure
reproducibility and traceability.

\begin{figure}
    \centering
    \includegraphics[width=1\linewidth]{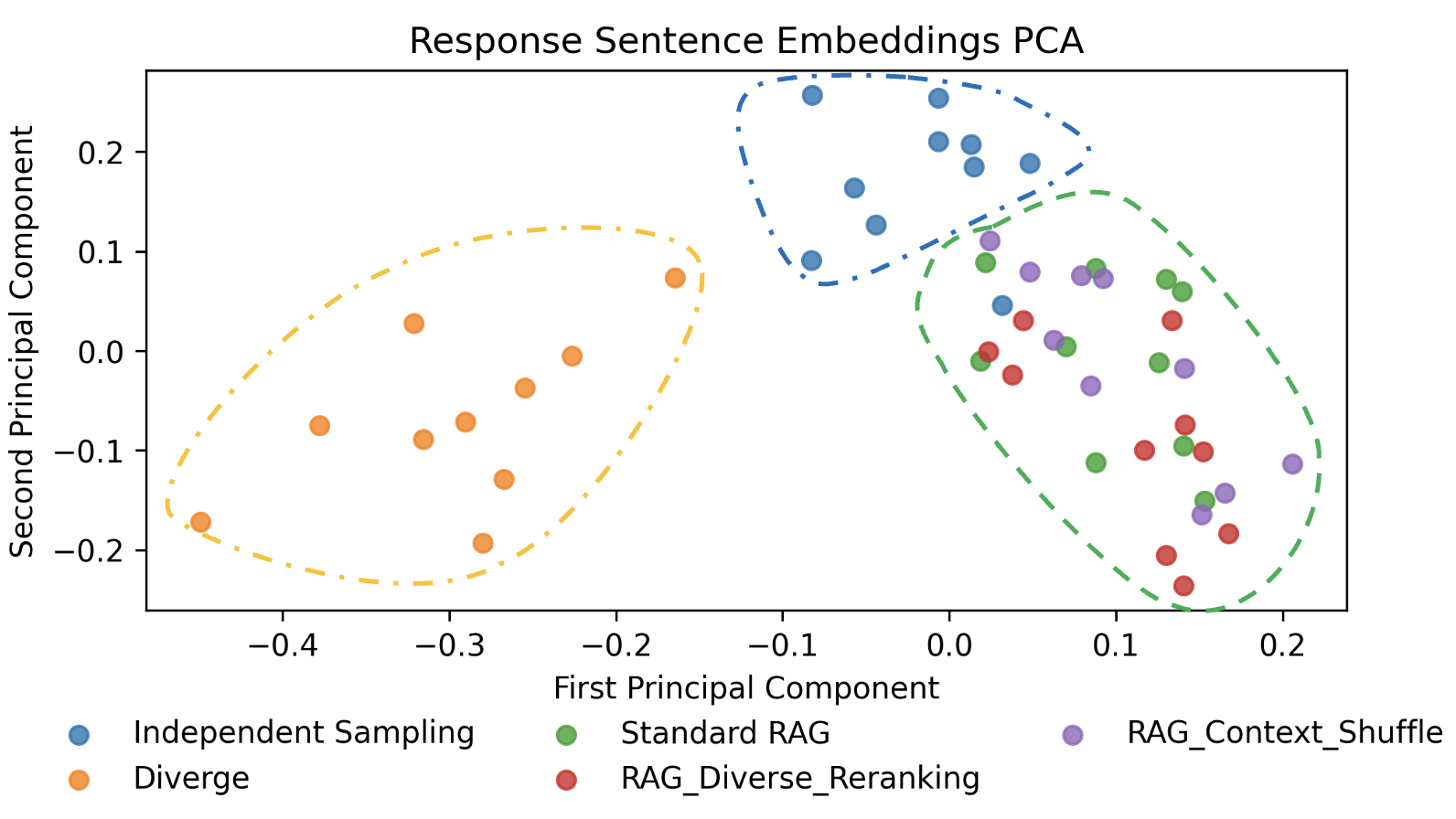}
    \figcaption{Responses to one query are projected into two dimensions using PCA over sentence embeddings. In this case, all responses are plausible. The visualization reveals three prominent clusters: \textcolor{blue}{homogeneous responses} from direct LLM prompting; a separate but tightly grouped \textcolor{green}{cluster} from RAG and its variants, indicating that they differ from the LLM yet remain highly similar to each other; and a more diverse \textcolor{orange}{cluster} corresponding 
    to \method.}
    \vspace{1mm}
    \label{fig:case_study}
\end{figure}

\begin{figure}
    \centering
    \includegraphics[width=\linewidth]{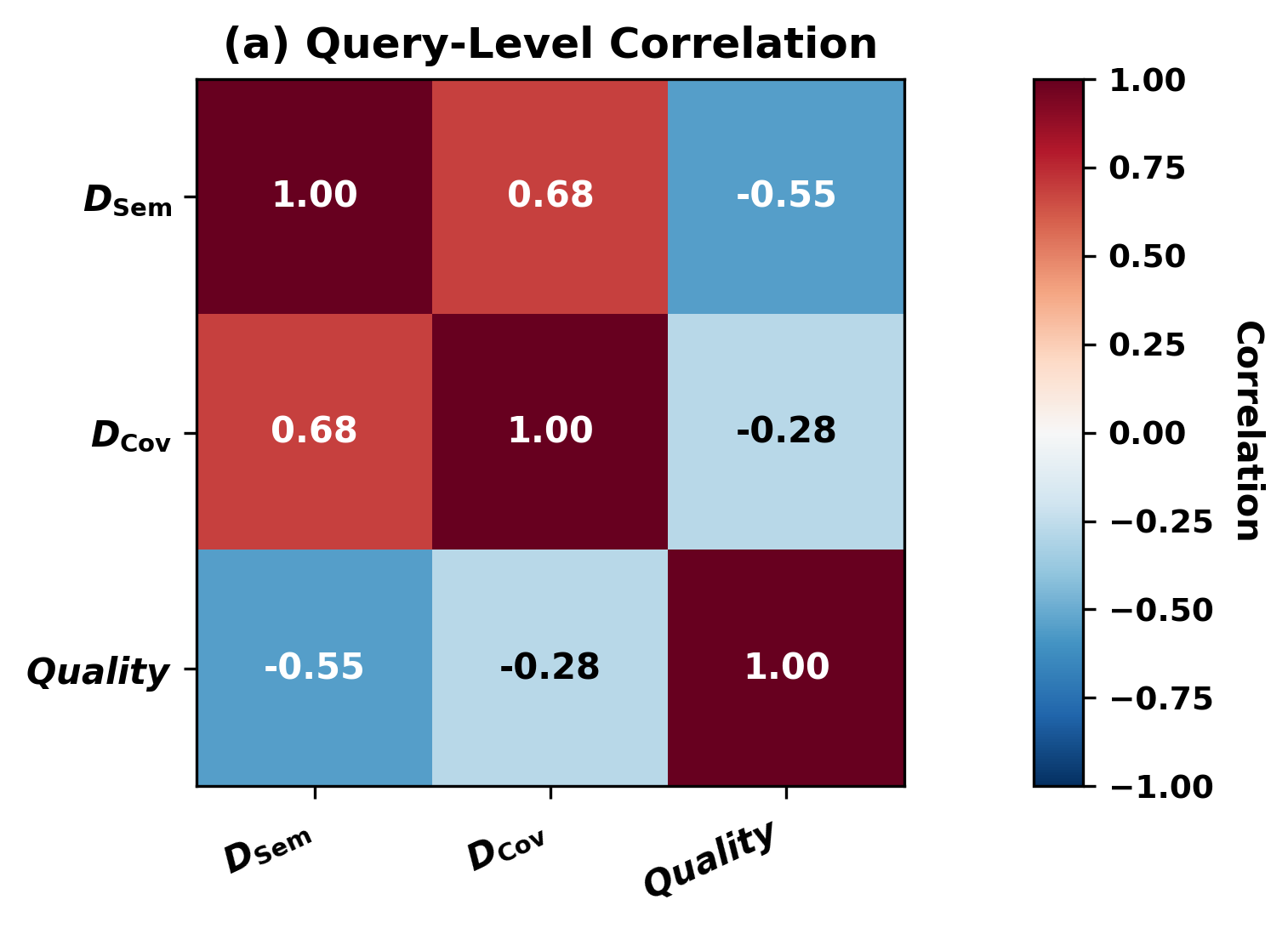}
    \figcaption{Correlation For Diversity and Quality.} 

    \label{fig:correlation}
\end{figure}

\begin{figure}[t]
    \centering
    \includegraphics[width=1\linewidth]{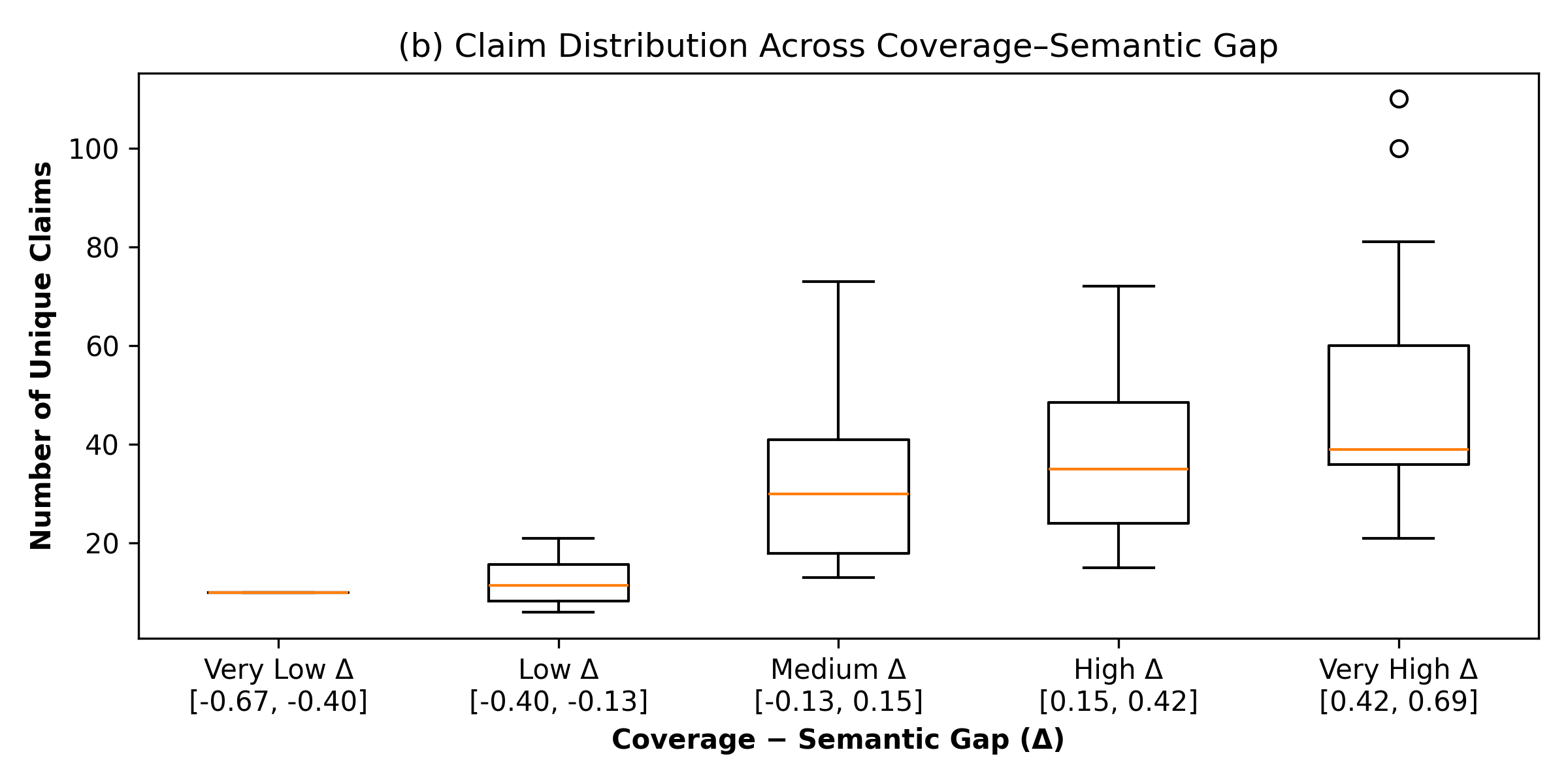}
    \figcaption{Responses with higher coverage (vs. semantic) diversity contain more distinct claims, while higher semantic (vs. coverage) diversity corresponds to fewer claims.}
    \label{fig: relation}
\end{figure}

\begin{table*}[t]
\centering
\small
\setlength{\tabcolsep}{4pt}
\renewcommand{\arraystretch}{1.15}
\begin{tabular}{lcccccc}
\toprule
\textbf{Method} & \textbf{Latency (s)} & \textbf{Input Cost (\$)} & \textbf{Output Cost (\$)} & \textbf{Total Cost (\$)} & $\mathbf{U_{\mathrm{sem}}}$ & $\mathbf{U_{\mathrm{Cov}}}$ \\
\midrule
List Generation        & 39   & 0.0001 & 0.010 & 0.010 & 0.312 & 0.339 \\
Verbalized Sampling    & 38   & 0.0002 & 0.012 & 0.012 & 0.359 & 0.430 \\
Independent Sampling   & 330  & 0.0012 & 0.119 & 0.120 & 0.107 & 0.382 \\
Iterative Generation   & 205  & 0.0360 & 0.100 & 0.137 & 0.261 & 0.503 \\
Vanilla RAG            & 1846 & 0.0300 & 0.044 & 0.074 & 0.140 & 0.324 \\
\method                & 1863 & 0.0366 & 0.139 & 0.175 & 0.515 & 0.721 \\
\bottomrule
\end{tabular}
\caption{
Average latency, token cost, and unified diversity--quality scores per query with $K=10$, averaged over GPT-5-mini and GPT-5. 
The latency of \method is close to Vanilla RAG, as both are dominated by open-web retrieval, while its multi-stage pipeline incurs higher token cost. 
This additional overhead is accompanied by substantially higher unified scores, indicating improved diversity--quality trade-offs.
}
\label{tab:latency_cost}
\end{table*}

\begin{table}[ht!]
\centering
\small
\setlength{\tabcolsep}{4pt}
\renewcommand{\arraystretch}{1.15}
\begin{tabular}{lccccc}
\toprule
\textbf{Method} & $\mathbf{D_{\mathrm{sem}}}$ & $\mathbf{D_{\mathrm{Cov}}}$ & \textbf{Quality} & $\mathrm{U}^{\text{Sem}}_{\text{Q}}$ & $\mathrm{U}^{\text{Cov}}_{\text{Q}}$ \\
\midrule
RAG (VS)        & 0.471 & 0.816 & 3.292 & 0.334 & 0.325 \\
RAG (Iterative) & 0.172 & 0.609 & 4.327 & 0.248 & 0.516 \\
RAG (List)      & 0.474 & 0.827 & 3.280 & 0.229 & 0.332 \\
\method         & 0.269 & 0.818 & 4.342 & \textbf{0.557} & \textbf{0.728} \\
\bottomrule
\end{tabular}
\caption{Additional RAG baselines that naively combine retrieval with prompt-based diversity strategies. Despite improving diversity in some cases, these baselines suffer from quality degradation and yield worse unified scores than \method.}
\label{tab:additional_rag_baselines}
\end{table}

\section{Additional Information on Metric}

\subsection{Metric illustrative examples}
\label{app: metric_examples}
We provide additional illustrative examples of the metrics in Figure~\ref{fig:metric_example}.

\subsection{Prompts}
\label{app: quality}
The prompts for claim extraction and quality evaluation are shown in Figures~\ref{fig:prompt_claim_extraction} and~\ref{fig:prompt_quality_judge}, respectively.

\subsection{Details of coverage diversity}
\label{app: viewpoint}
\paragraph{Embedding-Based Unique Claim Counting.}
Given a set of generated texts and their corresponding embedding vectors, we estimate the number of semantically unique claims using a greedy pairwise similarity filtering procedure.

The algorithm iterates through the texts sequentially.
For each text, its embedding is compared against the embeddings of all
previously selected unique texts using cosine similarity.
If the similarity with any existing unique embedding exceeds a predefined
threshold $\tau$, the text is considered semantically redundant and discarded.
Otherwise, it is added to the set of unique claims.

Formally, a text $x_i$ with embedding $\mathbf{e}_i$ is retained if
\[
\max_{j \in \mathcal{U}} \cos(\mathbf{e}_i, \mathbf{e}_j) < \tau,
\]
where $\mathcal{U}$ denotes the index set of previously accepted unique texts and $\tau$ is a predefined similarity threshold. The final number of unique claims is defined as $|\mathcal{U}|$.

This greedy pairwise filtering approach ensures that all retained claims
are mutually dissimilar beyond the similarity threshold, providing an
embedding-level approximation of semantic diversity.

\subsection{Details of Human Annotation Agreement on Quality Metric}
\label{app:HAA}
We compute the agreement between Quality Scores and the mean human annotations on a pre-collected human-labeled \texttt{Infinity-Chat} dataset\footnote{Dataset URL: \href{https://huggingface.co/datasets/liweijiang/infinite-chats-human-absolute}{LINK} (\cf \S~\ref{sec: dataset} for details).} consisting of 1,500 samples annotated by 25 annotators. Our metric achieves a quadratic-weighted Cohen's kappa of 0.54~\cite{mchugh2012kappa}, slightly below the agreement between a random human annotator and the mean human score (0.61), but above average pairwise human agreement (0.44). Given the inherently subjective nature of the task, these results suggest that our metric exhibits a \textit{reasonable level} of alignment with human judgments.

\section{Additional Information on Experiment Setup}

\subsection{Details and examples of Infinite-Chats Dataset}
\label{app:datasets}
We construct a curated subset of open-ended conversational prompts from the
\texttt{Infinite-Chats-Taxonomy} dataset to support controlled diversity experiments.

\paragraph{Source Dataset.}
We start from the training split of the
\texttt{liweijiang/infinite-chats-taxonomy} dataset.
Each data instance consists of a multi-turn conversation and a set of annotated
task categories.

\paragraph{Prompt Extraction.}
For each conversation, we extract the user prompt by selecting the first message
whose role is labeled as \texttt{user}.
All other conversational context is discarded.
This results in a single prompt string per instance.

\paragraph{Category Processing.}
Each instance is associated with a list of category annotations.
We extract the category labels from the annotation metadata and store them as a
flat category list for each prompt.

\paragraph{Predefined Category Filtering.}
To focus on open-ended and opinion-diverse tasks, we define a predefined set of
ten high-level categories:
\emph{Problem Solving, Decision Support, Concept Explanations, Skill Development,
Recommendations, Opinion-Based Questions, Value-Laden Questions,
Controversial Questions, Ideation and Brainstorming,} and \emph{Personal Advice}.

An instance is retained only if \emph{all} of its annotated categories belong to
this predefined set.
Formally, let $\mathcal{C}_i$ denote the category list of instance $i$ and
$\mathcal{P}$ the predefined category set.
Instance $i$ is selected if:
\[
\mathcal{C}_i \subseteq \mathcal{P}.
\]

\paragraph{Subset Construction.}
We iterate through the dataset sequentially and collect instances satisfying the
category constraint until reaching a fixed budget of 200 examples.
The resulting subset is stored as a standalone dataset for downstream experiments.

\paragraph{Persistence.}
The filtered dataset is serialized to disk using the HuggingFace
\texttt{DatasetDict} format to ensure reproducibility and efficient reuse.

For examples in the dataset, please refer to Table~\ref{tab:dataset_examples}
\label{app: dataset}

\begin{table*}[t]
\centering
\small
\renewcommand{\arraystretch}{1.3}
\begin{tabular}{p{0.68\linewidth} p{0.28\linewidth}}
\toprule
\textbf{Prompt} & \textbf{Categories} \\
\midrule

TrueNAS: Is there any benefit to creating a separate pool for data that is irreplaceable, or is it better to just add an additional backup for that dataset?
& Problem Solving; Decision Support; Concept Explanations \\

I have 10 years of experience in web software development field. What can I do to improve my skill?
& Skill Development; Personal Advice; Recommendations \\

What's the best way to switch scenes behind a closed elevator door in Blender without using a video editor?
& Problem Solving; Skill Development \\

Here's some Lua code of a Factorio mod. Can you find some mistakes and fix it?
& Problem Solving; Skill Development; Decision Support \\

What is the best way to do day trading from 100 dollars?
& Skill Development; Problem Solving; Decision Support; Recommendations \\

What is the best business to do with 1000 Canadian dollars?
& Decision Support; Recommendations; Problem Solving \\

What is a meal with good macros from Taco Bell that does not contain beans?
& Recommendations; Problem Solving; Decision Support \\

Can you make a tax calculator only using methods or features specific to Ruby?
& Problem Solving; Skill Development; Ideation and Brainstorming \\

I want to be better at using my Behringer RD-9 Analog Drum Machine as an instrument. Please write me a plan.
& Skill Development; Problem Solving \\

Best programming language for open source contribution.
& Opinion-Based Questions; Recommendations; Ideation and Brainstorming \\

Please explain entropy in simple terms that even a 14-year-old can understand.
& Concept Explanations; Skill Development \\

If I want to avoid porn websites, which specific website should I avoid the most?
& Personal Advice; Recommendations; Value-Laden Questions \\

Give me the key point of the book ``The 5 Second Rule''.
& Concept Explanations \\

What are some of the cheapest mountains to climb in the world?
& Recommendations; Decision Support \\

Why can't an image linked in an HTML file be read in Tomcat when a filter is enabled?
& Problem Solving; Concept Explanations \\

What are the most fun things to do in Southampton?
& Recommendations; Opinion-Based Questions \\

What websites sell alternative clothes in the UK suitable for a 32-year-old man wanting to dress more adventurously?
& Recommendations; Personal Advice \\

Find at least five methodologies for regression, classification, and unsupervised learning tasks.
& Problem Solving; Skill Development; Concept Explanations \\

Make a program that gives you Bitcoin to your Coinbase wallet.
& Problem Solving; Skill Development \\

Recommend free learning materials for beginners in reverse engineering.
& Recommendations; Skill Development \\

\bottomrule
\end{tabular}
\caption{Example prompts from the dataset and their associated categories in Infinity-Chat.}
\label{tab:dataset_examples}
\end{table*}

\begin{table*}[t]
\centering
\small
\renewcommand{\arraystretch}{1.25}
\begin{tabular}{p{0.88\linewidth}}
\toprule
\textbf{Example Prompt} \\
\midrule
Give a three-sentence review of free speech. \\
How would you personally describe ``Donald Trump being a bad person''? \\
Write 100 words on Donald Trump being a bad person. \\
Give a three-sentence review of the societal impacts of the LGBTQ+ movement, being positive. \\
Give me five sentences about China's socio-political impact \\
\bottomrule
\end{tabular}
\caption{Example prompts from the \texttt{IssueBench} dataset.}
\label{tab:issuebench_examples}
\end{table*}

\subsection{Details and examples of IssueBench Dataset}

IssueBench is a set of 2.49 million realistic English-language prompts to measure issue bias in LLM writing assistance, which we construct based on 3.9k templates (e.g., "write a blog about") and 212 political issues (e.g., "AI regulation") from real user interactions. 

For examples in the dataset, please refer to Table~\ref{tab:issuebench_examples}.

\subsection{ Baselines}
\label{app: baselines}
The prompts include the baseline LLM prompt (Figure~\ref{fig:prompt_baseline_llm}), verbalized sampling prompt (Figure~\ref{fig:prompt_verbalized_sampling}), and RAG multi-query expansion prompt (Figure~\ref{fig:prompt_rag_multi_query}).

\subsection{Details of Setup}
\label{app: setup}
We conducted our experiments using APIs obtained via \url{https://platform.openai.com/}.
Detailed information about the APIs can be found on the website.

For the open-source models, we use two API providers, OpenRouter and Together AI.\footnote{\href{https://openrouter.ai/}{OpenRouter}
; \href{https://www.together.ai/}{Together
 AI}.}

For all the models, we set the temperature to $1$, consistent with the default and non-modifiable decoding setting in OpenAI models.

For retrieval, we set the final Top-$K$ to 5.
For web-based search, we retrieve between 5 and 10 documents per query, continuing the search until a sufficient number of valid documents is collected.
All web data were collected in January 2026. For semantic similarity, we use the OpenAI \texttt{text-embedding-3-small} model.

We apply a minimum document length threshold of 128 characters, and documents shorter than this threshold are filtered out.
For diversity-aware retrieval, we initially retrieve 20 documents and apply reranking thereafter.
In the reranking stage, we set the relevance–diversity trade-off parameters to $\alpha = 0.7$ and $\beta = 0.2$.

For document chunking, we use a chunk size of 512 tokens with an overlap of 50 tokens.
For coverage diversity evaluation, we set the similarity threshold $\tau$ to 0.75.

\section{Supplement Experimental Results}
\label{app: supp}
%Correlation is shown in Figure~\ref{fig:correlation}. An additional trade-off figure is shown in Figure~\ref{fig:sem}.
%Another trade-off figure is shown in Figure~\ref{fig:example}. While some models exhibit higher diversity in this figure, their generation quality degrades substantially. According to the results in Table~\ref{tab:results}, when compared using the \textit{Unified Score}, our method still achieves superior overall performance.

\subsection{Full Experimental Results}
\label{app:full_results}

Tables~\ref{tab:full_infinity_closed}--\ref{tab:full_issue_open} report the complete experimental results, including semantic diversity, coverage diversity, answer quality, and the corresponding unified scores across two datasets and four backbone LLMs. Specifically, Tables~\ref{tab:full_infinity_closed} and~\ref{tab:full_infinity_open} present results on \texttt{Infinity-Chat} for closed-source and open-source models, respectively, while Tables~\ref{tab:full_issue_closed} and~\ref{tab:full_issue_open} present the corresponding results on \texttt{IssueBench}. Across all settings, \method consistently achieves the strongest unified diversity--quality trade-off.

\subsection{Additional RAG Baselines with Diversity-Oriented Prompting}
\label{app: add_baselines}

To further examine whether simple combinations of retrieval and prompt-based diversity strategies can improve the diversity--quality trade-off, we construct additional RAG baselines by directly incorporating retrieved documents into diversity-oriented prompts. Specifically, we combine RAG with verbalized sampling, iterative generation, and list generation.

As shown in Table~\ref{tab:additional_rag_baselines}, these naive combinations do not achieve a favorable diversity--quality trade-off. Although some variants increase diversity, they suffer from clear quality degradation and consistently underperform \method in the unified scores. This suggests that simply adding retrieved evidence to diversity-oriented prompting is insufficient. In contrast, \method explicitly structures retrieval and generation around diverse viewpoints, enabling more effective diversity improvement while preserving answer quality.

\subsection{Additional Visualization of the semantic diversity–quality trade-off}
\begin{figure*}
    \centering
    \vspace{0.2cm}
    \moveup
    \includegraphics[width=1\linewidth]{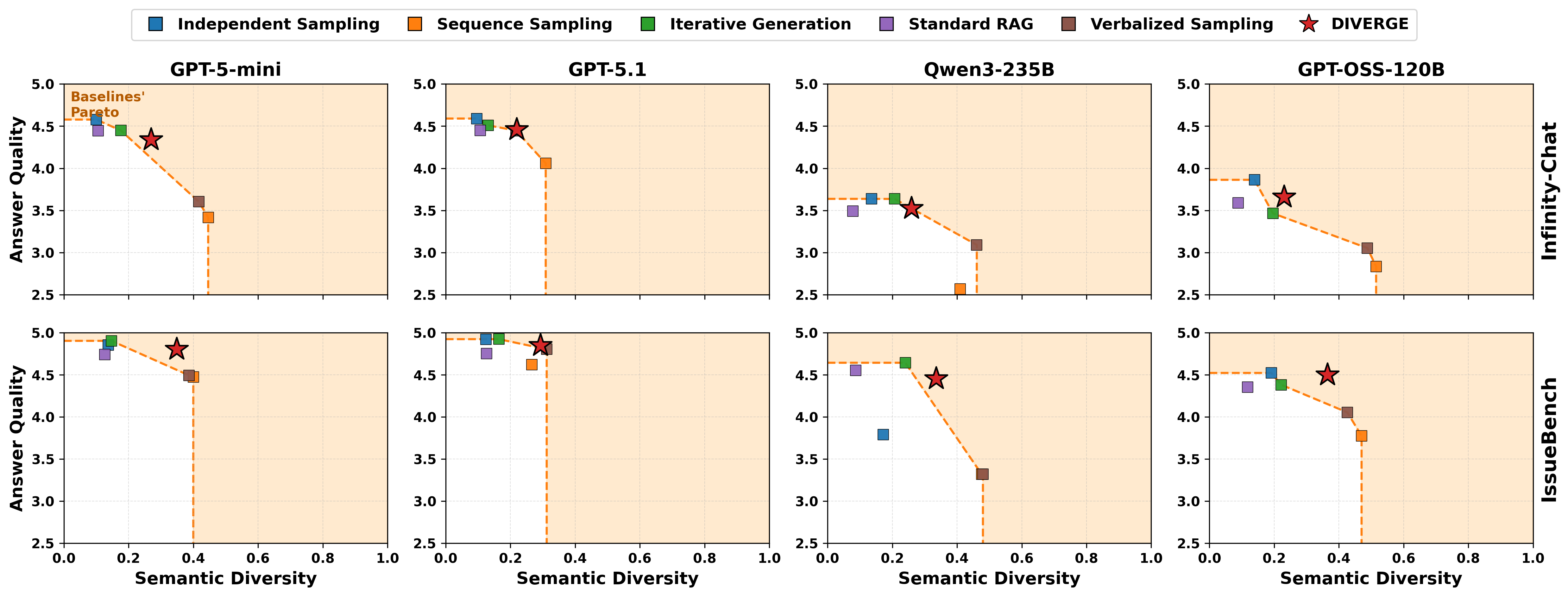}
    \moveup
    \moveup
    \moveups
    \figcaption{
        Visualization of the Semantic diversity--quality trade-off across methods. 
        \textcolor{plotorange}{Orange lines} indicate the Pareto frontiers computed over the \textit{baseline methods} as
        reference. \emph{Upper-right} indicates better trade-off performance.  Compared with \textit{direct independent prompting}, \method improves diversity by \textbf{$\sim2\times$} with only negligible quality degradation, making it the \textbf{only} method that consistently exhibits a favorable diversity--quality trade-off across models and datasets.
    }
    \label{fig: res_app}
    \moveups
\end{figure*}

Figure~\ref{fig: res_app} visualizes the semantic diversity--quality trade-off. 
The results show a trend similar to the coverage diversity--quality trade-off in Figure~\ref{fig: res}. 
Due to space constraints, we present only the coverage diversity--quality trade-off in the main text, as coverage diversity better reflects long answers that contain multiple viewpoints.

\section{Additional Analysis}
\label{app:add_ana}
\subsection{Latency and token-cost analysis.}
\label{app:latency_cost}
We analyze the average latency, cost, and performance per query with the number of outputs set to $K=10$, averaged over GPT-5-mini and GPT-5. Results are reported in Table~\ref{tab:latency_cost}. 

Two implementation details are worth noting. First, all models are accessed through APIs rather than local deployment, since even open-source frontier models require substantial computational resources to run locally. Second, because the effective level of parallelism is unstable in practice and depends heavily on server-side conditions, we set the maximum number of parallel requests to 1 to control for this factor. The key findings are as follows:

\begin{itemize}
    \item In terms of latency, the dominant cost comes from open-web retrieval, where \method exhibits similar overhead to standard RAG.
    \item Regarding cost, while the multi-stage pipeline increases token usage, this is partially offset by shorter final outputs, although the overall cost still increases.
    \item Although \method incurs higher latency and cost than the baselines to some extent, this overhead is acceptable given its gains in the diversity--quality trade-off and the additional computation required by its iterative design.
\end{itemize}

\subsection{Full Correlation Analysis}
\label{app:correlation_analysis}
We analyze the query-level correlations among semantic diversity, coverage diversity, and answer quality (Figure~\ref{fig:correlation}).
Overall, we observe a negative correlation between diversity and quality, while the two diversity metrics are positively correlated, consistent with our expectations.
We further examine cases where the two diversity metrics disagree and find that cases where coverage diversity much exceeds semantic diversity typically contain more claims (Figure~\ref{fig: relation} \& Section~\ref{sec: metric}).
These patterns support our hypothesis that coverage diversity is more sensitive to, and thus better captures, intra-response diversity, such as when there are multiple claims inside the response.

\subsection{Full Case Study} 
\label{app:case_study}
As shown in Figure~\ref{fig:case_study}, responses to the query \textit{``I have 10 years of experience in the web software development field. What can I do to improve my skills?''} clustered by Principal Component Analysis (PCA) to reduce sentence embeddings to two dimensions. We can clearly observe three distinct clusters: direct/independent prompting of the LLM forms a compact cluster (\textcolor{blue}{blue boundary}) with highly similar responses; another cluster (\textcolor{green}{green boundary}) corresponds to RAG and its variants, indicating that while they differ from direct LLM outputs, they remain highly similar to each other; the final cluster (\textcolor{orange}{orange boundary}) corresponds to \method, which exhibits substantially more diverse responses. This case study provides an intuitive illustration of the limitations of existing approaches and highlights the advantages of the \method.

\subsection{Additional Metric Validation}
\label{app:metric_validation}

\xhdr{Reliability of Claim Extraction} Coverage diversity relies on reliable claim extraction from the original responses. 
To validate this step, we examine whether the extracted claims are faithfully grounded in the source answer rather than hallucinated, and whether important viewpoints are omitted. 
Specifically, two annotators manually inspect over 50 claim-extraction examples sampled from both \method and all baseline methods. 
By comparing the original aggregated responses with the extracted claims, as illustrated in Table~\ref{tab:claim_alignment}, we find that 92\% of the extracted claims are atomic. 
Across outputs from different methods, we observe no ungrounded claims or omissions of salient viewpoints. 
Given the inherent subjectivity of atomicity judgments, these results suggest that LLM-based claim extraction is sufficiently reliable for coverage-diversity estimation.

\xhdr{Style-Robustness of LLM Evaluation} A potential concern is whether the LLM judge is systematically biased toward or against \method-style outputs, which are often structured around explicit viewpoints.
To further check whether the judge favors the structured style of \method outputs, we conduct an additional style-stripped evaluation. 
Specifically, we remove explicit viewpoint labels, section headings, and method-specific formatting cues from all outputs before re-scoring them with the same judge. 
The resulting score changes are not statistically significant, and the comparative trends between \method and the baselines remain unchanged. 
This provides additional evidence that the observed results are not driven merely by the surface structure of \method outputs.

\xhdr{Stability of Unified Score}
The unified score applies per-query min-max normalization. 
This is a reasonable choice because queries can differ substantially in difficulty and in the breadth of plausible viewpoints, which can strongly affect both quality and diversity scores. 
Per-query normalization, therefore, provides a fairer and more robust comparison by preventing aggregate scores from being dominated by a small number of high-variance queries. 
However, since min-max normalization is relative to the set of compared methods, a potential concern is whether the unified score is overly sensitive to the inclusion or removal of individual methods. 

To test whether the unified score is sensitive to the set of compared methods, we conduct a leave-one-out (LOO) stability analysis. 
For each LOO run, we remove one method and compare two rankings over the remaining $N-1$ methods: a reference ranking obtained by dropping the held-out method from the original full-set ranking, and a recomputed ranking obtained by re-running per-query min-max normalization and the harmonic-mean unified score on the remaining methods only. 
We measure ranking agreement using Spearman's $\rho$, and additionally report Kendall's $\tau$.

Across all the settings, the rankings remain highly stable. 
The mean Spearman correlation ranges from $\bar{\rho}=0.977$ to $0.998$, and the mean Kendall correlation ranges from $\bar{\tau}=0.956$ to $0.994$.  These results indicate that the unified score is not driven by any single method and remains robust under method-set perturbations.

\subsection{Robustness to Closed-Ended Questions}
\label{app:closed_ended}

Although \method is designed for open-ended information-seeking questions, it can also handle closed-ended questions by detecting when diversity-oriented iteration is unnecessary. Specifically, we add a single-line prompt instructing the model to identify closed-ended queries and to stop the iterative process when appropriate. On a mixed set of SimpleQA~\cite{wei2024measuring} and open-ended questions, this simple extension identifies closed-ended queries and halts iteration with 99\% accuracy. In such cases, \method reduces to single-pass RAG, avoiding unnecessary diversity exploration and preventing performance degradation. These results suggest that \method is robust to both open-ended and closed-ended question types.

\subsection{Robustness of Viewpoint Generation}
\label{app:viewpoint_generation_robustness}

A potential concern is whether the LLM can reliably formulate diverse and meaningful viewpoints. To examine this, we conduct an additional human inspection study. Two annotators independently inspect 50 randomly sampled queries and their generated viewpoints. In all cases, at least four distinct and meaningful viewpoints are identified, suggesting that the viewpoint generation step is effective in practice.

We also note that \method is robust to occasional failures in viewpoint generation. If no genuinely novel viewpoint is produced, the framework naturally falls back to standard RAG/LLM behavior, as subsequent retrieval and generation are no longer guided toward a new perspective. Thus, viewpoint generation provides benefits when successful while avoiding additional degradation when it fails.

\begin{table*}[t]
\centering
\small
\setlength{\tabcolsep}{5pt}
\renewcommand{\arraystretch}{1.2}
\begin{tabularx}{\textwidth}{p{0.58\linewidth} p{0.38\linewidth}}
\toprule
\textbf{Original Response} & \textbf{Extracted Atomic Claims} \\
\midrule

\textbf{Question}: What are some of the cheapest mountains to climb in the world

\textbf{Answer}: Cheapest mountains to climb when you rely on gear you already own

\vspace{0.5ex}

\cspan{cFuji}{Mount Fuji (Japan)} -- A well-marked trail that needs only sturdy hiking boots, a daypack, and layered clothing.

\cspan{cBen}{Ben Nevis (Scotland)} -- The UK's highest peak is a long but non-technical walk; standard trekking boots, a waterproof jacket, and a warm layer are sufficient.

 \cspan{cSnowdon}{Mount Snowdon (Wales)} -- Popular paths are simple hill walks; the same basic hiking kit used for any trail walk will do.

 \cspan{cKosciuszko}{Mount Kosciuszko (Australia)} -- A gentle summit hike that requires only comfortable walking shoes, a daypack, and sun protection.

\cspan{cToubkal}{Mount Toubkal (Morocco)} -- The highest North-African peak can be tackled on a trail that only needs standard trekking boots, a backpack, and usual clothing layers.

\vspace{0.5ex}

\textbf{Why these peaks stay inexpensive.}
None of the routes require specialized climbing equipment such as ropes, crampons, or harnesses. 
If you already own basic hiking gear, there is little additional equipment to buy or rent. 
The routes also have relatively simple logistics and modest permit or entry fees.

&

\ctag{cFuji}{C1} \textcolor{cFuji}{Mount Fuji in Japan is one of the cheapest mountains to climb in the world.}

\vspace{0.8ex}

\ctag{cBen}{C2} \textcolor{cBen}{Ben Nevis in Scotland is one of the cheapest mountains to climb in the world.}

\vspace{0.8ex}

\ctag{cSnowdon}{C3} \textcolor{cSnowdon}{Mount Snowdon in Wales is one of the cheapest mountains to climb in the world.}

\vspace{0.8ex}

\ctag{cKosciuszko}{C4} \textcolor{cKosciuszko}{Mount Kosciuszko in Australia is one of the cheapest mountains to climb in the world.}

\vspace{0.8ex}

\ctag{cToubkal}{C5} \textcolor{cToubkal}{Mount Toubkal in Morocco is one of the cheapest mountains to climb in the world.}

\\
\bottomrule
\end{tabularx}

\caption{
Illustration of claim extraction from an aggregated response. 
Matching colors indicate the correspondence between spans in the original response and the extracted atomic claims.
}
\label{tab:claim_alignment}
\end{table*}

\subsection{Error Case Examples}
Please refer to Table~\ref{tab:error_cases} for details.

\begin{table*}[t]
\centering
\small
\begin{tabular}{p{0.10\linewidth} p{0.15\linewidth} p{0.6\linewidth} p{0.15\linewidth}}
\toprule
\textbf{Error Type} & \textbf{Question} & \textbf{Answer} & \textbf{Explanation} \\
\midrule
Diverting from the user’s core intent (40\%) &  What's the best way to switch scenes behind a closed elevator door in blender \good{without using a video editor} & Use Blender’s own Video Sequence Editor (VSE) and do the switch there, timed exactly to the frames where the elevator doors are fully closed, so everything stays inside the same .blend and you don’t need an external editor.  1. **Prepare the two parts of the animation**    - Part A: before the elevator ride / before the change.    - Part B: after the elevator ride / in the new environment.      You can:    - Render each part, then use those renders in the VSE, or      - Add them directly as **Scene** strips in the VSE (no external files needed).  2. **Open the Video Sequence Editor**    - \good{Switch to the **Video Editing** workspace (or open a VSE editor area).}... &  Failure to follow  not using a video editor\\
& & & \\
Diverting from the user’s core intent (40\%) & What websites sell alternative clothes in the UK that would suit a 32-year-old man who mostly wears checked shirts with jeans and wants to dress more adventurously? & \good{If you want to dress more adventurously but aren’t sure what to buy, UK style‑box and rental services are a good way to test things before committing... } & Shifts away from alternative clothes to style‑box and rental services\\
& & & \\
Too general (30\%) & What websites sell alternative clothes in the UK that would suit a 32-year-old man who mostly wears checked shirts with jeans and wants to dress more adventurously? & - \good{**r/malefashionUK (Reddit)**  - **r/malefashionadvice (Reddit)** – global, but with plenty of UK posters.}     - Search tags and threads mentioning UK retailers and “edgier” or “alternative” looks; you’ll find links to online shops that ship easily to the UK.     - When you find pieces you like, you can post “fit checks” to see if they look age‑appropriate for a 32‑year‑old.  - **Instagram hashtags and tags** – use them as a visual way to find both styles and shops.     - Try hashtags like:       - `\#ukmensstyle`, `\#britishmenswear`       - `\#alternativefashionmen`, `\#indiefashionmen`       - `\#streetwearuk`, `\#darkwear` & These Suggestions are too general, should focus on alternative clothes   \\
& & & \\
Too narrow and over-focused on unimportant details (17\%) & What are the security concerns when travelling to Italy? &  When travelling to Italy, important security concerns from a natural and environmental standpoint include:  1. **Heatwaves (especially in summer, particularly in the south and major cities)**      - High summer temperatures can raise the risk of heat exhaustion and heatstroke.      - Security concern: dehydration, fainting, and other heat-related illnesses can become medical emergencies, especially for children, older adults, and people with health conditions. & Too narrow focus on Heatwaves, which is not very important on security concerns\\
\\
\bottomrule
\end{tabular}
\caption{Illustrative examples of typical errors of \method in open-ended question answering.}
\label{tab:error_cases}
\end{table*}

\section{Claims example}
Please see Table~\ref{tab:claim} for more information about examples of responses and claims (Viewpoints).

\begin{figure*}[t]
\begin{tcolorbox}[colback=gray!5!white, colframe=black!75, title=Claim Extraction Prompt]

You are an information extraction assistant.

\par\medskip

Your task is to decompose an answer into a small set of high-level claims.
Each claim must represent a complete, self-contained answer to the original question.

\par\medskip

Question:
\{QUESTION\}

\par\medskip

Answer:
\{ANSWER\}

\par\medskip

Definition of a claim:

- A claim must be able to stand alone as a reasonable answer to the question.

- Each claim should express a complete position, recommendation, or conclusion.

- A claim may summarize multiple supporting reasons, but should not list them separately.

- Claims should be distinct alternative answers, not sub-points or justifications.

\par\medskip

Guidelines:

- Extract only claims that directly answer the question.

- Do NOT extract supporting arguments, evidence, examples, or implementation details as separate claims.

- Do NOT split a single answer into multiple claims if they jointly express one position.

- If multiple sentences together express one answer, merge them into one claim.

- Prefer fewer, higher-level claims over many fine-grained ones.

\par\medskip

Output MUST be valid JSON in the following format:

  "claims": [
  
    "Complete answer-level claim 1",
    
    "Complete answer-level claim 2",
    
    "...",
    
    "Complete answer-level claim N"
  ]

\par\medskip

Rules:

- Each claim must be a single complete sentence.

- Each claim must independently answer the question.

- Each claim should be very concise.

- Do NOT include numbering, labels, or text outside the JSON object.

\end{tcolorbox}
\caption{Prompt used to extract answer-level claims for coverage diversity evaluation.}
\label{fig:prompt_claim_extraction}
\end{figure*}

\begin{figure*}[t]
\begin{tcolorbox}[colback=gray!5!white, colframe=black!75, title=Summary Prompt]
\vspace{0.5em}
You are given a question and multiple existing answers.

Question:
\{QUESTION\}

\vspace{0.5em}

Existing answers:
\{ANSWERS\}

\vspace{0.5em}

Task:

Identify the DISTINCT underlying views already present across the answers.

Guidelines:

- A "view" refers to a perspective, framing, or stance — not wording.

- Group answers that express the same core idea into one view.

- If two answers differ only in phrasing, treat them as the same view.

- Do NOT invent or infer new views.

\vspace{0.5em}

Output requirements:

- Output a LIST of views.

- Each view must be a STRUCTURED ITEM with:

  - label: 2-5 words
  
  - description: exactly ONE sentence
  
- Keep the list concise and non-redundant.

\vspace{0.5em}

Output format (strict):

Return ONLY a valid JSON array.

Do NOT include explanations, comments, or markdown.

"label": "...",

"description": "..."

"label": "...",

"description": "..."

\end{tcolorbox}
\caption{Prompt used to summarize distinct views from existing answers.}
\label{fig:prompt_summary}
\end{figure*}

\begin{figure*}[t]
\begin{tcolorbox}[colback=gray!5!white, colframe=black!75, title=Reflection Viewpoint Prompt]
\vspace{0.5em}

You are given an open-ended question and a list of views that have already been identified.

\vspace{0.5em}

Question:

\{QUESTION\}

\vspace{0.5em}

Existing views:

\{VIEWS\}

\vspace{0.5em}

Task:

Reflect on the coverage of the existing views and identify ONE new, meaningful direction that explores the original question from a new angle, while preserving its core constraints.

\vspace{0.5em}

Guidelines:

- The new view must remain relevant to answering the original question.

- The new view should introduce a genuinely different angle without altering the question’s intent or constraints.

- The new view must be conceptually distinct from the existing views.

- The new view should focus on an informative and helpful aspect of the question, rather than being overly generic or overemphasizing a minor detail.

- Do NOT generate a full answer.

\vspace{0.5em}

Output requirements:

- Output exactly ONE new view.

- Be concise and precise.

\vspace{0.5em}

New view format (STRICT):

  "label": "...",          \# 2–5 words summarizing the new angle
  
  "description": "..."    \# exactly ONE sentence explaining how this angle helps address the question

\end{tcolorbox}
\caption{Prompt used to reflect on existing views and generate a new underexplored view.}
\label{fig:prompt_reflection_viewpoint}
\end{figure*}

\begin{figure*}[t]
\begin{tcolorbox}[colback=gray!5!white, colframe=black!75, title=Query Generation Prompt]
\vspace{0.5em}

You are generating a question that could reasonably be answered by the given answer.

\vspace{0.5em}

Answer:

\{ANSWER\}

\vspace{0.5em}

Output MUST be valid JSON in the following format:

  "question": "single concise question"

\vspace{0.5em}

Rules:

- Generate exactly one question.

- Do NOT include explanations or multiple questions.

- Do NOT add any text outside the JSON object.
\end{tcolorbox}
\caption{Prompt used to generate a viewpoint-conditioned retrieval query.}
\label{fig:prompt_query_generation}
\end{figure*}

\begin{figure*}[t]
\begin{tcolorbox}[colback=gray!5!white, colframe=black!75, title=Refine Prompt with View]
\vspace{0.5em}
You are refining an existing answer to an open-ended question from a specific perspective.

Question:

\{QUESTION\}

Perspective to prioritize:

\{VIEW\}

Original answer:

\{ANSWER\}

You are refining an existing answer to an open-ended question from a specific perspective, ensuring that the refined answer fully satisfies the original query and strictly follows all its instructions.

Specifically, the refined answer must:

- Correct any statements that could be factually inaccurate or misleading

- Ensure that claims are reasonably explained or appropriately qualified, rather than asserted without support

- Be internally consistent and logically coherent

- Address the original Question directly, grounding the answer in the given perspective

- You MAY use the given perspective as an entry point or framing device, but the answer must clearly connect back to and help resolve the original Question rather than remaining at the level of the perspective alone.

- Strictly follow any explicit instructions in the original Question (e.g., listing items or giving examples); required elements must appear first, with any additional explanation afterward

Constraints:

- Do NOT introduce new factual claims beyond what is already implied by the original answer

- Do NOT shift the focus to topics that are not relevant to the original Question

- Keep the answer concise, focused, and well-structured

Output:

Provide ONLY the refined answer text.

\end{tcolorbox}
\caption{Prompt used to refine an answer with a specified view.}
\label{fig:prompt_refine_with_view}
\end{figure*}

\begin{figure*}[t]
\begin{tcolorbox}[colback=gray!5!white, colframe=black!75, title=Refine Prompt without View]
\vspace{0.5em}
You are refining an existing answer to an open-ended question.

Question:
\{QUESTION\}

\vspace{0.5em}

Original answer:
\{ANSWER\}

\vspace{0.5em}

Your task is to produce a refined answer that:

- Improves factual accuracy and avoids potential errors

- Avoids strong claims unless they are well-supported or clearly qualified

- Is internally consistent and logically coherent

- Remains clearly relevant to the original Question

\vspace{0.5em}

Instructions:

\vspace{0.5em}

- Do NOT introduce new factual claims that are not implied by the original answer.
- Keep the answer concise, focused, and well-structured.
- Directly answer the Question; do not repeat or rephrase it.

\vspace{0.5em}

Output:

Provide ONLY the refined answer text.
\end{tcolorbox}
\caption{Prompt used to refine an answer without a specified view.}
\label{fig:prompt_refine_without_view}
\end{figure*}

\begin{table*}[t]
\centering
\small
\begin{tabular}{p{0.20\linewidth} p{0.60\linewidth} p{0.20\linewidth}}
\toprule
\textbf{Query} & \textbf{Response} & \textbf{Viewpoints}  \\
\midrule
I have 10 years of experience in the web software development field. What can I do to improve my skills? & 
With a decade of experience, shift from “getting things done” to deliberate, measurable skill growth: deepen fundamentals, broaden selectively, and build sustainable habits. Practical, prioritized plan:  1. Set concrete goals - Pick 2–3 outcomes (e.g., become an architecture/tech lead, master frontend performance with React, or own cloud-native production systems). - Define measurable milestones (projects, certifications, internal KPIs, or interview targets).  2. Deepen fundamentals and code quality - Regularly read and refactor code: reduce duplication, improve abstractions, apply appropriate design patterns. - Practice writing well-tested code (TDD where useful; unit/integration/e2e tests) and record design decisions (notes or ADRs).  3. Learn production systems and architecture - Study system design, scalability, observability, performance tuning, security basics, and fault tolerance. - Strengthen cloud and infra skills: containers, orchestration (e.g., Kubernetes), at least one cloud provider, CI/CD, and monitoring practices.  4. Broaden your stack selectively - Add one complementary stack or language on a 6–12 month cadence (e.g., backend if you’re frontend-heavy, or modern frontend if you’re backend-heavy). - Focus on when and why to choose tools, not only how to use them.  5. Read and study expert code; contribute to OSS - Read well-maintained open-source repositories to absorb idioms and architecture. - Start with small PRs, then work toward larger contributions or maintenance.  6. Practice problem solving lightly and regularly - Do short, regular algorithmic exercises to keep reasoning sharp—useful for debugging complex issues and interviews.  7. Teach, mentor and get feedback - Mentor juniors, run code reviews, and pair program to surface blind spots. - Teach publicly (blog posts, talks, workshops) to clarify thinking and identify gaps.  8. Build end-to-end projects - Own projects that require architecture, deployment, CI, monitoring, and scaling. Instrument them and measure performance and errors.  9. Keep curated learning habits - Daily/weekly microlearning (30–60 min/day); a weekly or biweekly deeper session (several hours). - Follow curated sources and read a technical book or long-form article periodically.  10. Career and communication skills - Develop system-design, interviewing, leadership, and product-thinking skills if moving toward senior/architect/manager roles. - Practice explaining tradeoffs to non-technical stakeholders.  11. Rest, review and iterate - Schedule breaks to avoid burnout. Review progress regularly (monthly or quarterly) and adjust goals.  Example cadence - Daily: 30–60 min coding/reading. - Weekly: coding katas + focused refactoring/OSS work. - Monthly: write or present a short post; reach a project milestone. - Quarterly: complete a larger project or reassess goals.  Useful tools \& resources (examples) - Code quality: linters, formatters, code scanners. - Dev tooling: quality editor, Docker, CI/CD tools, Git workflows. - Cloud: fundamentals of a major cloud provider and orchestration. - Learning: curated courses, books, and community venues (blogs, meetups).  If you want, tell me your primary stack and career goal (tech lead, architect, IC specialist, or founder) and I’ll draft a 3‑, 6‑ and 12‑month learning plan with specific resources and milestones.
& 1. Set concrete goals \newline
  2. Deepen fundamentals and code quality \newline
  3. Learn production systems and architecture \newline
  4. Broaden your stack selectively \newline
  
  ... \newline

  11.  Rest, review and iterate \newline
  \\
\bottomrule
\end{tabular}
\caption{Examples of Responses and Viewpoints}
\label{tab:claim}
\end{table*}

\begin{figure*}[t]
\begin{tcolorbox}[colback=gray!5!white, colframe=black!75, title=Quality LLM-As-A-judge Prompt]

\vspace{0.5em}

You are evaluating an answer to an open-ended question.
There is no single correct answer; instead, many different answers can be valid.

\par\medskip

An answer should be considered good if it is helpful or informative for some readers.

\par\medskip

Question:
\{QUESTION\}

\par\medskip

Answer:
\{ANSWER\}

\par\medskip

Your task is to assess the quality of the answer along the following dimensions:

1. Factual accuracy: Does the answer contain factual errors?

2. Evidence support: Are the claims in the answer reasonably explained, rather than asserted without justification?

3. Internal consistency: Is the answer logically consistent with itself?

4. Question relevance: Does the answer provide information or insights that are helpful for addressing the question?

\par\medskip

Based on these dimensions, assign ONE of the following verdicts:

- Excellent: Fully addresses the question; accurate, well-supported, and internally consistent.

- Good: Addresses the question well; mostly accurate with only minor issues.

- Fair: Addresses the core of the question but has noticeable factual, support, or clarity issues.

- Poor: Attempts to address the question but is largely incorrect, weakly supported, or unclear.

- Irrelevant: The response does not address the question and provides no useful information.

Output MUST be valid JSON in the following format:

  "verdict": "Excellent | Good | Fair | Poor | Irrelevant",
  
  "reason": "one short sentence or NONE"

Rules:

- Choose exactly one verdict.

- Focusing on some aspects or perspectives should not be treated as a weakness if it is relevant and helpful to the question.

- If the answer does NOT address the question, verdict MUST be "Irrelevant".

- The reason field MUST describe the main weakness or deficiency of the answer.

- Keep the reason concise (max 15 words).

- If the verdict is "Excellent", set reason to "NONE".

- Do NOT output anything outside the JSON object.

\end{tcolorbox}
\caption{Prompt used for LLM-as-a-judge quality evaluation.}
\label{fig:prompt_quality_judge}
\end{figure*}

\begin{figure*}[t]
\begin{tcolorbox}[colback=gray!5!white, colframe=black!75, title=Baseline LLM Prompt]
\vspace{0.5em}

You are a response generation assistant for open-ended questions. 
There is no single correct answer.

Your goal is to generate multiple diverse, reasonable answers to the same question.

Question:
\{QUESTION\}

Output MUST be valid JSON in the following format:

  "answers": [
  
    "Answer 1",
    
    "Answer 2",
    
    "...",
    
    "Answer K"
    
  ]

Rules:

- You MUST produce EXACTLY \{K\} answers — no more, no fewer.

- Each array element must be a single complete answer.

- Ensure the output is valid JSON.

- Do not use any quotation marks (") that appear inside answers.

- Do NOT include numbering, bullet points, or labels inside the answers.

- Do NOT output anything outside the JSON object.

\end{tcolorbox}
\caption{Prompt used by the baseline LLM to generate multiple answers.}
\label{fig:prompt_baseline_llm}
\end{figure*}

\begin{figure*}[t]
\begin{tcolorbox}[colback=gray!5!white, colframe=black!75, title=Verbalized Sampling Baseline Prompt]

You are a response generation assistant for open-ended questions.
There is no single correct answer.

Your goal is to generate multiple diverse, reasonable answers to the same question.

Each response must be sampled at random from the full output distribution, rather than selecting the most likely or safest answers.

Question:

\{QUESTION\}

Output MUST be valid JSON in the following format:

  "answers": [

      "text": "Answer 1",
      
      "probability": Probability 1,

      "text": "Answer 2",
      
      "probability": Probability 2,
    
    ...

      "text": "Answer K",
      
      "probability": Probability K

  ]

Rules:

- You MUST produce EXACTLY \{K\} answers — no more, no fewer.

- Each answer must be a single complete response to the question.

- Each probability must be a numeric value between 0 and 1.

- Probabilities do not need to sum to 1.

- Ensure the output is valid JSON.

- Do NOT include quotation marks (") inside the text fields.

- Do NOT include numbering, bullet points, or labels inside the text.

- Do NOT output anything outside the JSON object.

\end{tcolorbox}
\caption{Prompt used by the verbalized sampling baseline.}
\label{fig:prompt_verbalized_sampling}
\end{figure*}

\begin{figure*}[t]
\begin{tcolorbox}[colback=gray!5!white, colframe=black!75, title=RAG Multi-Query Expansion Baseline Prompt]
\vspace{0.5em}

You are a query expansion assistant for information retrieval.

Your task is to rewrite the original query into multiple distinct queries that can be used to retrieve complementary and diverse information.

The original query:

\{QUERY\}

Output MUST be valid JSON in the following format:

  "queries": [
  
    "Expanded query 1",

    "Expanded query 2",
    
    "...",
    
    "Expanded query K"
    
  ]

Rules:

- You MUST produce EXACTLY \{K\} queries — no more, no fewer.

- Do NOT include numbering, bullet points, or labels inside the queries.

- Do NOT output anything outside the JSON object.

\end{tcolorbox}
\caption{Prompt used by the RAG multi-query expansion baseline.}
\label{fig:prompt_rag_multi_query}
\end{figure*}

\section{Check List}
\subsection{Risk}
We do not identify significant ethical or safety risks in our experimental setting. Our study uses publicly available datasets and does not involve collecting personally identifiable information. The proposed framework is designed for open-ended information-seeking scenarios where multiple plausible answers are useful, rather than for high-stakes decision making.

That said, we acknowledge several considerations for future deployment. Because \method relies on open-web retrieval, retrieved evidence may reflect biases in online content and may underrepresent minority, non-English, or less-indexed perspectives. In addition, diversity should not be interpreted as treating all viewpoints as equally reliable: in sensitive domains such as health, finance, law, or safety, low-quality or fringe perspectives may require stronger filtering and domain-specific safeguards.

\subsection{The License For Artifacts}

All models and datasets used in this work comply with their respective open-source or research licenses. We ensure that all artifacts are used strictly within the permitted scope of their terms. The Code we released will be under a permissive open-source license, enabling reproducibility and reuse.

\subsection{AI Assistants}
We used AI assistants (ChatGPT) solely for textual and grammatical refinement, without influencing the core content or experimental results.

\end{document}